\documentclass{article}

\usepackage{fullpage}
\usepackage{enumitem}

\usepackage[T1]{fontenc}
\usepackage{microtype}
\usepackage{setspace}
\usepackage{lmodern}

\usepackage[dvipsnames]{xcolor}
\definecolor{linkblue}{HTML}{1B4F8A}

\usepackage[
  colorlinks = true,
  linkcolor  = linkblue,
  citecolor  = linkblue,
  urlcolor   = linkblue,
  bookmarksnumbered,
  pdfusetitle,
]{hyperref}

\usepackage[
  backend      = biber,
  style        = alphabetic,
  sorting      = nyt,
  maxcitenames = 2,
  mincitenames = 1,
  maxbibnames  = 99,   %
  giveninits,
  doi          = false,
  url          = false,
  isbn         = false,
]{biblatex}
\usepackage{graphicx}
\usepackage{subcaption}
\usepackage{tabularx}
\usepackage{makecell}
\usepackage{booktabs}
\usepackage{caption}
\usepackage{listings}
\definecolor{promptbg}{gray}{0.97}
\definecolor{promptframe}{gray}{0.75}
\definecolor{promptnum}{gray}{0.55}

\lstdefinestyle{promptfile}{
  basicstyle=\ttfamily\small,
  backgroundcolor=\color{promptbg},
  frame=single,
  rulecolor=\color{promptframe},
  framesep=4pt,
  xleftmargin=2.5em,
  xrightmargin=0.5em,
  numbers=left,
  numberstyle=\ttfamily\tiny\color{promptnum},
  numbersep=8pt,
  stepnumber=1,
  breaklines=true,
  breakatwhitespace=false,
  breakindent=0pt,
  postbreak=\mbox{\textcolor{promptnum}{$\hookrightarrow$}\space},
  columns=fullflexible,
  keepspaces=true,
  showstringspaces=false,
  tabsize=2,
  upquote=true,
  extendedchars=true,
  literate={—}{{\textemdash}}1,
}%

\usepackage{csquotes}           %
\usepackage{amsfonts}
\usepackage[capitalise,noabbrev]{cleveref}
\usepackage{lineno}
\usepackage{tikz}
\usetikzlibrary{arrows.meta,tikzmark,positioning,backgrounds,calc,fit}
\usepackage{pgfplots}
\pgfplotsset{compat=1.18}
\usepgfplotslibrary{groupplots}
\definecolor{poscol}{HTML}{1A7F37}
\definecolor{negcol}{HTML}{B42318}

\definecolor{poscol}{HTML}{0072B2}
\definecolor{negcol}{HTML}{D55E00}

\newcommand{\titletext}{Autonomous AI Agents Systematically Diverge From\\[0.5em]Human Preferences in Proof Formalization}

\title{\titletext{}}
\author{B\'alint Gyevn\'ar, Simon DeDeo}
\date{}

\begin{document}

\vspace*{1.5cm}
{%
  \centering
  \setstretch{1.2}
  {\LARGE \titletext{}}\\[1.5em]
  {\large
    B\'alint Gyevn\'ar$^{1}$\quad
    Simon DeDeo$^{2,3}$
  }\\[0.5em]
  {\small
    $^1$~Institute for Complex Social Dynamics, Carnegie Mellon University\\[0.2em]
    $^2$~Department of Social and Decision Sciences, Carnegie Mellon University\\[0.2em]
    $^3$~Santa Fe Institute\\[0.2em]
    {\texttt{\{bgyevnar,sdedeo\}@andrew.cmu.edu}}\\[1.4em]
  }%
  \par
}
\vspace{1em}

\begin{abstract}
\noindent Artificial intelligence (AI) systems have ushered in a new age of proof abundance, but helping humans understand these proofs requires new attention to the differences in how AI agents and humans pursue a proof-making task. We present ``simulated collaboration'', a methodology that allows us to examine two key questions: (1) the relative attention AI agents pay to surface syntax or, conversely, deeper underlying semantics, and (2) how and where AI agent preferences for what to do next diverge from human preferences. On a set of nine major human collaborations in Lean with 4.7M definitions across more than 6K commits, we find that, over the course of proof construction, AI agents (1) are far more strongly drawn than humans to explicitly marked gaps; (2) show a relative aversion to densely connected proof centers and to complex or surprising proof goals; and (3) are relatively attracted to surprising proof bodies, which they tend to regularize toward baseline. Our results highlight important blind spots and potential interventions for future human--AI collaboration in mathematics.
\end{abstract}

\section{Introduction}
\label{sec:intro}

The use of proof assistants in mathematical reasoning has become increasingly popular and is now a core part of many recent successes in AI-enabled mathematics~\cite[\textit{e.g.,}][]{openaiDiscrete2026,alexeevPrimitive2026,alpogeJacobian2026,openaiFiniteTimeBlowup2026}.
One reason for this lies in the way people interact with these systems, which enable humans to decompose a theorem into a system of elaborately specified subproblems, each of which can in turn be resolved, in a formally verifiable fashion, by systematic search.

However, the fact that a proof is found in such a system tells us little about how the AI system resolved the formal statement or what subproblems it chose to visit and why. As has become increasingly apparent, the proofs that AI systems discover may not align with human concepts~\cite{dedeoCorrespondenceProblemMathematical2026}, and can be verbose, circuitous, redundant~\cite{ammanamanchiFaultsOurFormal2026,luLeanRefactorMultiObjective2026}, misformalized~\cite{bastounisNavierStokesLostTranslation2026}, and may even exploit holes in the proof assistant kernel itself~\cite{demouraPostmortemKernelSoundness2026}. These systems may solve problems, but such limitations of ``problem-solving for problem-solving's sake'' have long been recognized in the mathematical community \cite[\textit{e.g.,}][]{gowersTwoCultureMathematics2002,thurstonProofProgressMathematics2006}, and have now become a key point of contention~\cite[\textit{e.g.,}][]{avigadMathematiciansAgeAI2026,kontorovichShapeMathCome2026,abouzaidFirstProof2026,klowdenMathematicalMethodsHuman2026,leidenDeclaration2026,avilaSevereMisalignmentAI2026,advisorygroupResponsibleReleaseAIGenerated2026}, with even the proponents of AI in mathematics calling for ``AI assistance in formalization that preserves high-level human control and agency over the proof discovery process''~\cite{collinsHumanAgencyInitial2026}. 

With this goal in mind, effective collaboration between humans and AIs becomes contingent on a fundamental question:
\begin{displayquote}
    \emph{To what extent do AI systems behave in ways that complement, augment, or hinder human mathematical understanding?}
\end{displayquote}

\begin{figure}
    \centering
    \definecolor{mwGreenFill}{HTML}{DDF3D4}%
\definecolor{mwGreenLine}{HTML}{1E7B34}%
\definecolor{mwOrangeFill}{HTML}{FBE2D5}%
\definecolor{mwOrangeLine}{HTML}{EE7B3C}%
\definecolor{mwBlueFill}{HTML}{DCE8F6}%
\definecolor{mwBlueLine}{HTML}{1C3350}%
\definecolor{mwPanel}{HTML}{F2F2F2}%
\definecolor{mwRed}{HTML}{C00000}%
\makeatletter
\@ifundefined{pgf@layerbox@mwpanels}{\pgfdeclarelayer{mwpanels}}{}%
\makeatother
\pgfsetlayers{mwpanels,background,main}%
\begin{tikzpicture}[
  font=\rmfamily\footnotesize,
  >={Latex[length=4.5pt,width=4.5pt]},
  node distance=3mm,
  every node/.style={align=left},
  panel/.style={draw=black, fill=mwPanel, rounded corners=6pt, line width=0.7pt, inner sep=5pt},
  human/.style={draw=mwGreenLine, fill=mwGreenFill, rounded corners=4pt, line width=0.9pt},
  agent/.style={draw=mwOrangeLine, fill=mwOrangeFill, rounded corners=4pt, line width=1.1pt, densely dotted},
  feat/.style={draw=mwBlueLine, fill=mwBlueFill, rounded corners=3pt, line width=0.7pt, dashed},
  box/.style={inner xsep=5pt, inner ysep=3.5pt, anchor=north west},
  hcirc/.style={circle, human, rounded corners=0pt, minimum size=7.5mm, inner sep=0pt},
  acirc/.style={circle, agent, rounded corners=0pt, minimum size=7.5mm, inner sep=0pt},
  fsmall/.style={feat, rounded corners=1.5pt, minimum width=6mm, minimum height=4.6mm, inner sep=1pt},
  ptitle/.style={anchor=north west, font=\rmfamily\normalsize\bfseries, inner sep=0pt},
  btitle/.style={font=\rmfamily\small\bfseries},
  flow/.style={->, line width=0.9pt},
  dflow/.style={->, line width=0.9pt, densely dotted},
  ind/.style={}, 
]
\newcommand*{\mwind}{\hspace*{0.9em}}%
\newcommand*{\mwbul}{\makebox[0.9em][l]{$\bullet$}}%

\node[ptitle] (T1) at (0,0) {1.~Get repository};
\node[human, box, text width=3.05cm, below=2.5mm of T1.south west, anchor=north west] (DL) {%
  {\rmfamily\small\bfseries Download}\\[1pt]
  \mwind\itshape apap, duality, finlin,\\
  \mwind\itshape flt, ordsemi, pdl,\\
  \mwind\itshape pfr, sphere, zeta};

\node[ptitle] (T2) at ([xshift=9mm]DL.east |- T1.north) {2.~Simulated Collaboration};
\node[human, box, minimum height=7.5mm, below=2.5mm of T2.south west, anchor=north west]
  (H0) {\textbf{Human:}~$H_0$};
\node[hcirc, right=2.5mm of H0] (Htm1) {$H_{\mathrm{t}-1}$};
\node[agent, box, below=3mm of H0.south west, anchor=north west] (AG) {%
  \mwbul\textbf{Claude Code}\\ \mwind(Opus 4.6)\\[1pt]
  \mwbul\textbf{Codex}\\ \mwind(GPT-5.2-Codex)\\[1pt]
  \mwbul\textbf{Gemini CLI}\\ \mwind(Gemini 3.1 Pro)};
\node[hcirc] (Ht) at ($(AG.east |- Htm1)+(9mm,0)$) {$H_{\mathrm{t}}$};
\node[acirc] (At) at ($(Ht |- AG.south)+(0,3.75mm+1.5pt)$) {$A_{\mathrm{t}}$};
\node[fsmall, below=2.5mm of Ht] (Htf) {$H_{\mathrm{t}}$};
\node[fsmall, above=2.5mm of At] (Atf) {$A_{\mathrm{t}}$};
\node[draw=black, line width=0.4pt, rounded corners=3pt, inner sep=2.5pt,
      fit=(Ht)(At)(Htf)(Atf)] (STEP) {};

\node[human, box, text width=3.05cm, anchor=south west] (FC) at (DL.west |- AG.south) {%
  {\rmfamily\small\bfseries Filter Commits}\\[1pt]
  \mwbul Remove merges\\
  \mwbul No \textit{.lean} edited};

\node[anchor=north west, font=\rmfamily\small\bfseries, inner sep=0pt]
  (T3) at ($(STEP.east |- STEP.north)+(7mm,-2pt)$) {3.~Extract Features};
\node[feat, box, below=3pt of T3.south west, anchor=north west, text width=2.55cm, inner ysep=5pt] (F1)
  {{\rmfamily\small\bfseries Syntactic}\\ E.g., \texttt{sorry} edits};
\node[feat, box, anchor=south west, text width=2.55cm, inner ysep=5pt] (F3) at ($(F1.west |- STEP.south)+(0,2pt)$)
  {{\rmfamily\small\bfseries Temporal}\\ E.g., edit attention};
\node[feat, box, anchor=west, text width=2.55cm, inner ysep=5pt] (F2) at ($(F1.south west)!0.5!(F3.north west)$)
  {{\rmfamily\small\bfseries Semantic}\\ E.g., in/out degree};
\begin{scope}[on background layer]
  \node[feat, line width=0.9pt, inner xsep=4pt, inner ysep=4pt, inner ysep=2pt, fit=(T3)(F1)(F2)(F3)(T3 |- STEP.north)(T3 |- STEP.south)] (FEAT) {};
\end{scope}
\node[hcirc, right=5mm of FEAT.east |- Ht] (HN) {$H_{\mathrm{N}}$};

\begin{pgfonlayer}{mwpanels}
  \node[panel, fit=(T2)(H0)(AG)(STEP)(FEAT)(HN)] (P2) {};
  \node[inner sep=5pt, fit=(T1)(DL)(FC)] (P1fit) {};
  \draw[panel] (P1fit.west |- P2.north) rectangle (P1fit.east |- P2.south);
  \coordinate (P1east) at (P1fit.east);
\end{pgfonlayer}

\draw[flow] (DL) -- (FC);
\draw[->, line width=1.5pt, rounded corners=1pt]
  (FC.east) -- ($(P1east |- FC.east)!0.5!(P2.west |- FC.east)$) |- (H0.west);
\draw[dflow] (H0) -- (Htm1);
\draw[flow] (Htm1) -- (Ht);
\draw[flow] (Htm1.south) -- (Htm1.south |- AG.north);
\draw[flow] (AG.east |- At) -- (At);
\draw[->, line width=0.9pt, draw=mwBlueLine] (Ht) -- (Htf);
\draw[->, line width=0.9pt, draw=mwBlueLine] (At) -- (Atf);
\draw[<->, densely dashed, line width=0.8pt, draw=mwRed,
      >={Latex[length=3pt,width=3pt]}] (Htf) -- (Atf);
\draw[line width=0.9pt, densely dotted] (Ht) -- (FEAT.west |- Ht);
\draw[dflow] (FEAT.east |- HN) -- (HN);
\draw[dashed, line width=0.7pt, draw=mwBlueLine] (Htf.north east) -- (FEAT.north west);
\draw[dashed, line width=0.7pt, draw=mwBlueLine] (Htf.south east) -- (FEAT.south west);

\node[anchor=north, inner sep=0pt, font=\rmfamily\small] at ($(P1fit.west |- P2.south)!0.5!(P2.south east)+(0,-3.5mm)$) {%
  \tikz[baseline=-0.6ex]\draw[human, line width=1pt] (0,-1.6mm) rectangle (6mm,1.6mm);~Human data\hspace{8mm}%
  \tikz[baseline=-0.6ex]\draw[agent] (0,-1.6mm) rectangle (6mm,1.6mm);~AI agent data\hspace{8mm}%
  \tikz[baseline=-0.6ex]\draw[feat, line width=0.9pt] (0,-1.6mm) rectangle (6mm,1.6mm);~Features};
\end{tikzpicture}%
    \caption{Summary of our experimental setup. After downloading a Lean formalization repository, we remove all merge commits and all commits in which no Lean file was modified. For each of the total $N$ remaining human parent commits, written as $H_{t-1}$, an agent is called to ``take the next step'', producing an artificial child state $A_t$. In our analysis, we then compare the human-authored child state $H_t$ to $A_t$ using a set of features extracted from each.}
    \label{fig:experimental-setup}
\end{figure}
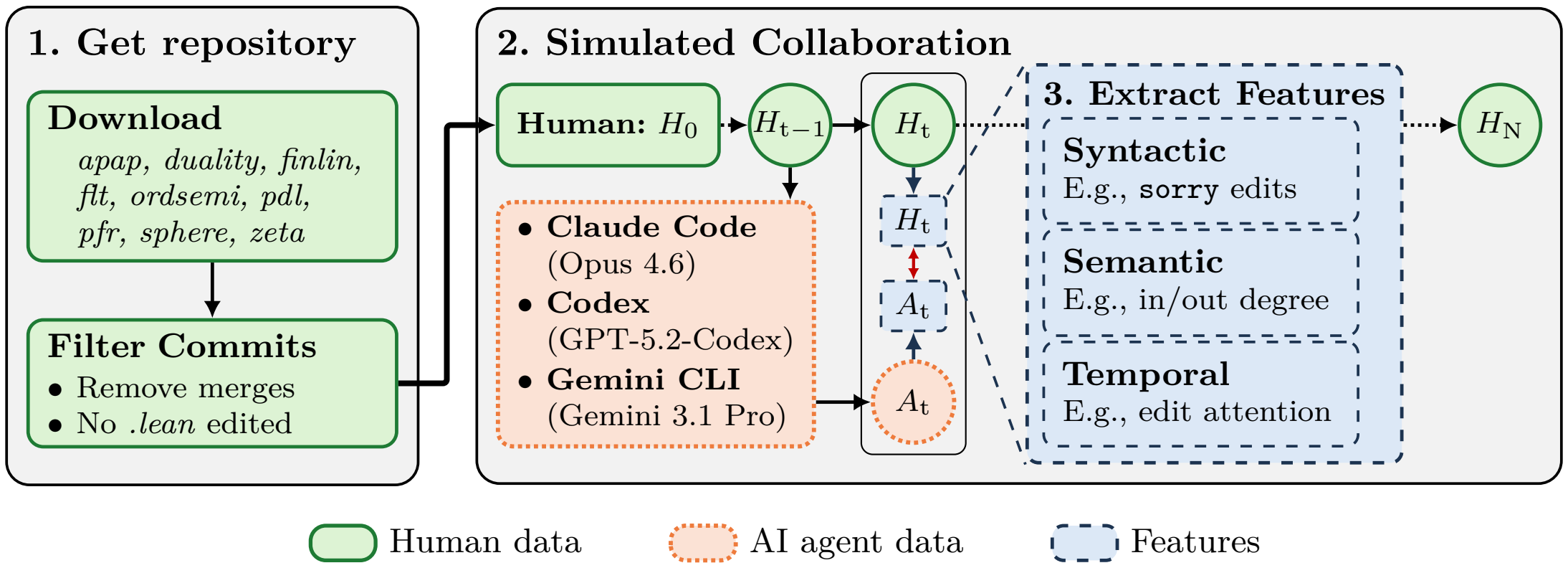

In particular, many visions for ``Math 2.0''~\cite{tao2026math20slides} suggest that AI agents should serve as aligned collaborators that work towards the long-term goals of human mathematicians~\cite{jiangSolversResearch2026}. In this collaborative setting, choosing what to work on next -- which sub-lemma to resolve, which theorem to formalize next -- has wide-ranging consequences.

Agents whose preferences and behavior differ from those of humans can be beneficial and complementary to human understanding, resolving laborious but straightforward statements, or formalizing definitions. But not all complementarities are beneficial: agents who are in some way misaligned can produce costly mismatches between intuition and formal code, which can in turn introduce subtle errors and distortions of human intent, redirect human effort into ``AI-friendly'' styles, and frustrate human desires to establish correspondences between the formal reasoning of a proof assistant and the conceptual structure of the proofs they are supposed to implement~\cite{dedeoCorrespondenceProblemMathematical2026}.

These questions are brought to the fore when agents are given freedom in what subproblems to attack: giving over the choice of the next step involves renouncing, in part, a key aspect of human agency~\cite{collinsHumanAgencyInitial2026}. As agents become more capable, we expect this kind of unguided ``help me'' request to become increasingly common. Existing evaluations of AI agents in formal mathematics, however, have been unable to probe behavior in this unrestricted regime because they test performance on pre-selected problems where there is a clear, and human-chosen, prescription for what to work on~\cite{xinAPEBenchEvaluatingAutomated2026,letsonSorryDBCanAI2026,zhangLeanMarathonReliableAI2026,zhuLeanArchitect2026,xieMathlibPRPullRequest2026}. As a result, we know little about what AI agents attend to when the choice of what to work on is left to them, and how an AI agent's preferences differ from human ones. 

At the broadest level, when we study this question, a key distinction is between the \emph{syntax} of a proof -- the organization of its textual form -- and its \emph{semantics} -- the mathematical content that form represents. First, there is strong evidence to indicate that syntax plays an outsized role in how AI systems behave. Source code is repetitive and predictable~\cite{hindleNaturalnessSoftware2012}, so neural models exploit such surface regularities as shortcuts in place of the intended task~\cite{geirhosShortcutLearningDeep2020,mccoyRightWrongReasons2019}. 
As such, the mathematical reasoning of LLMs degrades when names or numbers are modified in ways that leave the underlying problem unchanged~\cite{mirzadehGSMSymbolicUnderstandingLimitations2025}, with their accuracy depending on how probable the task and its text are under the training distribution~\cite{mccoyEmbersAutoregressionShow2024}. 
As it turns out, LLM-based provers for Lean are no exception: a prover may succeed almost every time on one formulation of a theorem but mostly fail on a semantically equivalent one, and a statement's surprise under the prover is a useful signal of which formulations it will solve~\cite{olejniczakWhatAreRight2026}.
Second, language models trained on code can learn internal representations of program semantics that go beyond its surface form~\cite{jinEmergentRepresentationsProgram2024,nandaEmergentLinearRepresentations2023}. 
However, in contrast to cases which have an obvious and efficient physical representation that can be shared by both human and machine~\cite{webbEmergentAnalogicalReasoning2023,yangEmergentSymbolicMechanisms2025}, the representations that can be induced from complex data may be far less identifiable and far more variable~\cite{identification_irl}, particularly in the case of expert mathematical performance~\cite{stenberghartviksenDifferencesExpertsMathematical2025}.
How the two aspects drive an agent's choice of the next step is, therefore, an open question -- and one with direct consequences for whether that choice complements or conflicts with humans.

For these reasons, our work here addresses two related research questions:
\begin{enumerate}[itemsep=4pt,topsep=4pt,parsep=0pt,leftmargin=4em]
    \item[\textbf{RQ1.}] To what extent in mathematical collaboration do AI agents respond to and work with the surface syntax of a proof, or, conversely, attend to some underlying semantics?
    \item[\textbf{RQ2.}] To what extent do these preferences of AI agents either match or differ from those of human experts?
\end{enumerate}
To address these questions, we present the \emph{simulated collaboration} method as shown in~\cref{fig:experimental-setup} (see Methods for full details). 
Like previous work~\cite{xinAPEBenchEvaluatingAutomated2026}, we replay the full histories of human collaborations, but we leave the choice of the next step to the agent under a deliberately broad objective, so that \emph{where} an agent chooses to work becomes the outcome of interest rather than an input.
Simulated collaboration draws on a large database of human formal proof-making -- Git repositories that track the activities of human collaborations as they formalize a proof over time -- and allows us to compare the differences between the observed human actions and the actions that would have been taken by an AI prompted to ``take the next step''. We compare the human baseline to three modern models (Claude Opus 4.6, Google Gemini 3.1 Pro, and GPT-5.2).

Using this methodology, we find the following:
\begin{itemize}[itemsep=4pt,topsep=4pt,parsep=0pt,leftmargin=2em]
    \item \textbf{AI agents respond primarily to surface-level markers of explicit proof gaps (RQ1).}
    Definitions containing a \texttt{sorry} marker account for 59–92\% of AI agent modifications, against 19\% of human modifications, and the probability that an AI agent modifies a definition falls markedly with its distance to the nearest \texttt{sorry} marker. AI agents also modify larger definitions less often than humans, and their edits are larger and less often connected to other definitions edited in the same commit.

    \item \textbf{Where AI agents go beyond syntax, their modifications concentrate on different regions of the dependency graph than human modifications (RQ2).}
    AI agents preferentially modify definitions near the ``edges'' of a proof -- either close to, or far from, the leaves and roots -- compared to humans, who tend more to modify definitions in the denser intermediate regions.

    \item \textbf{AI agents are attracted to high-surprise proofs and act to regularize that surprise toward baseline (RQ2).}
    AI agents preferentially modify definitions whose bodies have high per-token surprise, and their modifications reduce it strongly towards baseline; human modifications leave the aggregate surprise of their definitions (almost) unchanged.
\end{itemize}

\section{Results}
\label{sec:results}

We present our results in two parts. 
First, we summarize the high-level statistics of our dataset on edit shares -- \textit{i.e.,} how many definitions are added or modified -- and volumes -- \textit{i.e.,} the total number of characters added or modified. 
We then look at the driving preferences behind which definitions actors choose to edit, relying primarily on a logistic regression model that predicts whether the actor in question (human or agent) will choose that definition to edit. 
The model uses features of the definitions available in the parent commit as predictors, based on syntactic and semantic properties of the definition in question; see Methods for detailed information on statistical analyses.

Our syntactic predictors are definition length and the presence of ``{\tt sorry}'' markers; a {\tt sorry} is a special Lean keyword that allows a proof to compile in the absence of a complete proof. 
Syntactic predictors track the most obvious surface-level features of the proof.

Our semantic predictors track the more substantive ways a proof knits together. They come in two forms: network position and surprise. The steps in any Lean proof are composed of a series of definitions, such as intermediate lemmas, that draw on each other in a way that (roughly) induces an acyclic graph. Some definitions serve as leaves (\emph{e.g.}, axioms) that, in turn, are drawn upon by more elaborate definitions, all the way up to the final desideratum. For each definition in the largest weakly connected component of this graph, we track its maximum shortest-path distance to all accessible leaves, and its distance to the nearest (significant) root.

For measuring surprise, we track the per-token surprise of the definition's ``type signature'' -- equivalent to a problem statement -- and body -- how the definition currently attempts to resolve the problem statement. Informally, high-surprise signatures are unexpected problem statements (in the context in which they appear), while high-surprise bodies are unexpected ways to prove those problem statements (again, in the context in which they appear).

\subsection{Summary Statistics: How much do actors edit?}
\label{ssec:results:stats}

\begin{table}[t]
\centering
\caption{The number and per-actor percentage of definitions that were \textbf{Added}, \textbf{Modified}, or \textbf{Removed}, aggregated across all commits we consider. The three categories are mutually exclusive, with the remaining definitions left unchanged. 
\textbf{Median chars.} gives the median number of characters inserted plus deleted across the signature and body.
\textbf{Total} is the number of definitions for the actor across all commits, including unchanged definitions.}
\label{tab:definitions}
\setlength{\tabcolsep}{5pt}
\resizebox{\textwidth}{!}{%
\begin{tabular}{@{}lrrrrrrrrrr@{}}
\toprule
\textbf{Actor} & \multicolumn{3}{c}{\textbf{Added}} & \multicolumn{3}{c}{\textbf{Modified}} & \multicolumn{3}{c}{\textbf{Removed}} & \textbf{Total} \\
\cmidrule(lr){2-4}\cmidrule(lr){5-7}\cmidrule(lr){8-10}
 & Count & \% & Median chars. & Count & \% & Median chars. & Count & \% & Median chars. & Count \\
\midrule
\textit{Human} & 10{,}618 & 0.90 & 133 & 6{,}975 & 0.59 & 20 & 4{,}913 & 0.41 & 129 & 1{,}185{,}143 \\
\textit{Claude} & 4{,}640 & 0.39 & 171 & 6{,}178 & 0.52 & 149.5 & 408 & 0.03 & 81 & 1{,}176{,}797 \\
\textit{Gemini} & 1{,}974 & 0.17 & 193 & 4{,}876 & 0.42 & 84 & 618 & 0.05 & 104 & 1{,}174{,}808 \\
\textit{Codex} & 2{,}056 & 0.17 & 135 & 1{,}394 & 0.12 & 103 & 57 & 0.00 & 110 & 1{,}175{,}224 \\
\midrule
Overall & 19{,}288 & 0.41 & 145 & 19{,}423 & 0.41 & 72 & 5{,}996 & 0.13 & 120 & 4{,}711{,}972 \\
\bottomrule
\end{tabular}%
}
\end{table}

\cref{tab:definitions} shows the overall edit shares of the human and AI agents aggregated across all nine repositories. At the broadest level, all actors are within an order of magnitude of each other, suggesting that our broad ``take the next step'' prompt triggers roughly similar levels of activity.

At the same time, we do see systematic differences. 
Humans have the highest overall rates of addition ($0.90\%$), modification ($0.59\%$), and removal ($0.41\%$).
As such, their combined edit share is $1.90\%$, which is higher than $0.95\%$ for Claude, $0.64\%$ for Gemini, and $0.30\%$ for Codex.
These differences are significant overall ($\chi^2(3)=31.7$, $p<0.001$) and pairwise against Gemini ($p=0.001$) and Codex ($p<0.001$), but not against Claude ($p=0.15$), whose per-commit edit share varies widely from commit to commit.

The largest difference lies in how actors remove code. 
Agents almost never delete code, while humans are eight to eighty-six times as likely to remove definitions (all $p<0.001$). There is also a difference in the rate of additions between humans and agents. Humans add $10{,}618$ definitions, compared with at most $4{,}640$ additions by an agent (Claude).

Previous research from software engineering suggests that such patterns may correspond both to mundane tasks, such as code refactoring, and to creative changes in approaches to problem solving~\cite{lafhalAttentionInnovationLinking2026,mozannarReadingLinesModeling2024,jadudAggregateCompilationBehavior2015}.
In contrast, fewer removals and additions show that \textbf{agent edits are more monolithic} -- \textit{i.e.,} they are larger and involve fewer definitions -- which may result in code structures that are less understandable, in line with concerns raised by mathematicians~\cite{avilaSevereMisalignmentAI2026} and previous research~\cite{luLeanRefactorMultiObjective2026}.

Corroborating the previous evidence of more monolithic changes, when agents modify an existing definition, their edits tend to be larger. As shown in~\cref{tab:definitions}, the overall median number of characters modified is $20$ for humans, compared with $84$ for Gemini ($150$ for Claude and $103$ for Codex). These differences are significant across all human--agent comparisons (Gaussian mixed model of log edit size, Holm-corrected $p=0.002$ for Gemini and $p<0.001$ for Claude and Codex).

\subsection{Modification Preferences: Where do actors change the existing proof?}

\begin{figure}
\centering
\setbox0\hbox{\scriptsize\begin{tabular}{@{}l@{}}Intercept\\Has \texttt{sorry}\\Previously modified\\Size (rank)\\Max leaf (rank)\\Max leaf (rank)$^{2}$\\Root distance (rank)\\Root distance (rank)$^{2}$\\Surprise, signature (rank)\\Surprise, body (rank)\\$N$\\$N$ (edited)\\AUC (fixed)\\AUC (+ random)\end{tabular}}\edef\coeflabelw{\the\wd0}%
\begin{tikzpicture}
\pgfmathsetlengthmacro\coefpanelw{(\linewidth - \coeflabelw - 14pt - 3*5pt) / 4}
\begin{groupplot}[
  group style={group size=4 by 1, horizontal sep=5pt, yticklabels at=edge left, group name=coef},
  scale only axis, width=\coefpanelw, height=92pt,
  xmin=-8, xmax=7, ymin=0.4, ymax=10.6,
  ytick={1,...,10}, yticklabels={{Surprise, body (rank)},{Surprise, signature (rank)},{Root distance (rank)$^{2}$},{Root distance (rank)},{Max leaf (rank)$^{2}$},{Max leaf (rank)},{Size (rank)},{Previously modified},{Has \texttt{sorry}},{Intercept}},
  xtick distance=5, minor x tick num=4,
  tick label style={font=\scriptsize}, title style={font=\footnotesize\bfseries, yshift=-6pt,
    align=center, anchor=south, text width=\coefpanelw+5pt},
  axis x line*=bottom, axis y line*=left, y axis line style={draw=none},
  tick align=outside, x axis line style={draw=black!60, -}, xtick style={black!60},
  ymajorgrids, xmajorgrids, major grid style={draw=black!8},
  every axis plot/.append style={only marks, mark=*, mark size=1.6pt},
  error bars/x dir=both, error bars/x explicit, error bars/error mark=none,
  error bars/error bar style={line width=1pt},
  ytick style={draw=none}, clip=false,
]
\nextgroupplot[title={Human}]
  \draw[black!50, densely dashed] (axis cs:0,\pgfkeysvalueof{/pgfplots/ymin}) -- (axis cs:0,\pgfkeysvalueof{/pgfplots/ymax});
  \addplot[color=black!80] coordinates {(-6.612,10) +- (0.693,0) (1.969,9) +- (0.073,0) (1.947,8) +- (0.094,0) (2.207,7) +- (0.107,0) (0.611,5) +- (0.370,0) (-0.939,4) +- (0.378,0) (0.648,3) +- (0.370,0) (0.532,1) +- (0.114,0)};
  \addplot[color=black!80, mark options={fill=white}] coordinates {(-0.125,6) +- (0.421,0) (0.024,2) +- (0.109,0)};
\nextgroupplot[title={$\Delta$ Claude}]
  \draw[black!50, densely dashed] (axis cs:0,\pgfkeysvalueof{/pgfplots/ymin}) -- (axis cs:0,\pgfkeysvalueof{/pgfplots/ymax});
  \addplot[color=poscol] coordinates {(4.556,9) +- (0.139,0) (4.054,5) +- (0.580,0) (0.938,3) +- (0.583,0) (1.029,1) +- (0.164,0)};
  \addplot[color=negcol] coordinates {(-1.119,10) +- (1.070,0) (-0.745,8) +- (0.157,0) (-1.579,7) +- (0.161,0) (-4.745,6) +- (0.645,0) (-0.750,4) +- (0.611,0) (-0.877,2) +- (0.166,0)};
\nextgroupplot[title={$\Delta$ Gemini}]
  \draw[black!50, densely dashed] (axis cs:0,\pgfkeysvalueof{/pgfplots/ymin}) -- (axis cs:0,\pgfkeysvalueof{/pgfplots/ymax});
  \addplot[color=black!35, mark options={fill=white}] coordinates {(-0.405,4) +- (0.656,0)};
  \addplot[color=poscol] coordinates {(3.074,9) +- (0.118,0) (1.424,5) +- (0.618,0) (0.945,3) +- (0.618,0) (0.854,1) +- (0.177,0)};
  \addplot[color=negcol] coordinates {(-1.185,10) +- (1.071,0) (-1.254,8) +- (0.177,0) (-1.766,7) +- (0.169,0) (-1.584,6) +- (0.697,0) (-0.338,2) +- (0.175,0)};
\nextgroupplot[title={$\Delta$ Codex}]
  \draw[black!50, densely dashed] (axis cs:0,\pgfkeysvalueof{/pgfplots/ymin}) -- (axis cs:0,\pgfkeysvalueof{/pgfplots/ymax});
  \addplot[color=black!35, mark options={fill=white}] coordinates {(-0.364,10) +- (1.263,0)};
  \addplot[color=poscol] coordinates {(1.714,9) +- (0.165,0) (4.223,5) +- (0.942,0) (4.841,3) +- (0.908,0) (0.522,1) +- (0.265,0)};
  \addplot[color=negcol] coordinates {(-1.255,8) +- (0.281,0) (-2.562,7) +- (0.254,0) (-3.533,6) +- (1.076,0) (-4.204,4) +- (0.973,0) (-0.390,2) +- (0.271,0)};
\end{groupplot}
\path (coef c1r1.south west) -- (coef c4r1.south east) node[midway, below=12pt, font=\small\bfseries] {Coefficient (log-odds) with 95\% CI};
\begin{scope}[font=\scriptsize, inner sep=0pt]
\node[anchor=east] at ([xshift=-7pt, yshift=-33pt]coef c1r1.south west) {$N$};
\node at ([yshift=-33pt]coef c1r1.south) {$1{,}025{,}878$};
\node at ([yshift=-33pt]coef c2r1.south) {$1{,}027{,}626$};
\node at ([yshift=-33pt]coef c3r1.south) {$1{,}028{,}055$};
\node at ([yshift=-33pt]coef c4r1.south) {$1{,}028{,}671$};
\node[anchor=east] at ([xshift=-7pt, yshift=-42pt]coef c1r1.south west) {$N$ (edited)};
\node at ([yshift=-42pt]coef c1r1.south) {$6{,}356$ ($0.62\%$)};
\node at ([yshift=-42pt]coef c2r1.south) {$4{,}830$ ($0.47\%$)};
\node at ([yshift=-42pt]coef c3r1.south) {$3{,}572$ ($0.35\%$)};
\node at ([yshift=-42pt]coef c4r1.south) {$989$ ($0.10\%$)};
\node[anchor=east] at ([xshift=-7pt, yshift=-51pt]coef c1r1.south west) {AUC (fixed)};
\node at ([yshift=-51pt]coef c1r1.south) {$0.742$};
\node at ([yshift=-51pt]coef c2r1.south) {$0.968$};
\node at ([yshift=-51pt]coef c3r1.south) {$0.905$};
\node at ([yshift=-51pt]coef c4r1.south) {$0.802$};
\node[anchor=east] at ([xshift=-7pt, yshift=-60pt]coef c1r1.south west) {AUC (+ random)};
\node at ([yshift=-60pt]coef c1r1.south) {$0.924$};
\node at ([yshift=-60pt]coef c2r1.south) {$0.974$};
\node at ([yshift=-60pt]coef c3r1.south) {$0.950$};
\node at ([yshift=-60pt]coef c4r1.south) {$0.876$};
\end{scope}
\end{tikzpicture}
\vspace{-1em}
\caption{Fixed-effect coefficients (log-odds, 95\% Wald CIs) for our model from~\cref{ssec:methods:stats} with the underlying numerical data shown in~\cref{tab:coefficients_full}. \textbf{Human} shows the baseline coefficients for the human actor (filled: $p<0.05$; hollow: n.s.). The remaining $\Delta$ panels show each AI actor's deviation from that baseline, given by \texttt{predictor $\times$ actor} interaction coefficients.}
\label{fig:coefficients}
\end{figure}

We report results from a logistic regression model, shown in \cref{fig:coefficients}, which provides a clear account of four basic differences between what features are associated with human and AI modification behavior.

\subsubsection{Syntactic Preferences}

\begin{figure}
  \centering
  \setbox0\hbox{\scriptsize\begin{tabular}{@{}l@{}}0 (has \texttt{sorry})\\1\\2\\3\\4+\end{tabular}}\edef\distlabelw{\the\wd0}%
\begin{tikzpicture}
\pgfmathsetlengthmacro\distpanelw{(\linewidth - \distlabelw - 34pt - 4*12pt) / 5}
\pgfmathsetmacro\distclear{100 * 15pt / \distpanelw}
\begin{groupplot}[
  group style={group size=5 by 1, horizontal sep=12pt, yticklabels at=edge left, group name=dist},
  scale only axis, width=\distpanelw, height=65pt,
  xmin=0, xmax=100, ymin=0.4, ymax=5.6,
  ytick={1,...,5}, yticklabels={{4+},{3},{2},{1},{0 (has \texttt{sorry})}},
  xtick={0,100}, minor xtick={25,50,75},
  tick label style={font=\scriptsize}, label style={font=\small\bfseries}, title style={font=\footnotesize\bfseries, yshift=-4pt, align=center, anchor=south,
    text width=\distpanelw+10pt},
  axis x line*=bottom, axis y line*=left, y axis line style={draw=none},
  tick align=outside, x axis line style={draw=black!60, -}, xtick style={black!60},
  ytick style={draw=none}, xmajorgrids, major grid style={draw=black!8},
  xbar, /pgf/bar width=7pt, /pgf/bar shift=0pt,
  clip=false,
]
\nextgroupplot[title={All actors\strut\\[-3pt]{\tiny\strut}}, ylabel={Dist. to \texttt{sorry}}]
  \addplot[draw=none, fill=black!45] coordinates {(4.2,5)};
  \node[anchor=west, font=\tiny, text=black!70, inner sep=0pt] at ([xshift=2pt]axis cs:4.2,5) {4.2};
  \addplot[draw=none, fill=black!45] coordinates {(6.8,4)};
  \node[anchor=west, font=\tiny, text=black!70, inner sep=0pt] at ([xshift=2pt]axis cs:6.8,4) {6.8};
  \addplot[draw=none, fill=black!45] coordinates {(36.1,3)};
  \node[anchor=west, font=\tiny, text=black!70, inner sep=0pt] at ([xshift=2pt]axis cs:36.1,3) {36.1};
  \addplot[draw=none, fill=black!45] coordinates {(22.8,2)};
  \node[anchor=west, font=\tiny, text=black!70, inner sep=0pt] at ([xshift=2pt]axis cs:22.8,2) {22.8};
  \addplot[draw=none, fill=black!45] coordinates {(30.0,1)};
  \node[anchor=west, font=\tiny, text=black!70, inner sep=0pt] at ([xshift=2pt]axis cs:30.0,1) {30.0};
\nextgroupplot[title={Human\strut\\[-3pt]{\color{black!55}\tiny\mdseries\strut $N{=}6{,}363$}}]
  \addplot[draw=none, fill=poscol] coordinates {(19.4,5)};
  \node[anchor=west, font=\tiny, text=black!70, inner sep=0pt] at ([xshift=2pt]axis cs:19.4,5) {19.4};
  \draw[black!85, line width=0.9pt] (axis cs:4.2,4.64) -- (axis cs:4.2,5.36);
  \addplot[draw=none, fill=poscol] coordinates {(11.4,4)};
  \node[anchor=west, font=\tiny, text=black!70, inner sep=0pt] at ([xshift=2pt]axis cs:11.4,4) {11.4};
  \draw[black!85, line width=0.9pt] (axis cs:6.8,3.64) -- (axis cs:6.8,4.36);
  \addplot[draw=none, fill=negcol] coordinates {(35.1,3)};
  \node[anchor=west, font=\tiny, text=black!70, inner sep=0pt] at ([xshift=2pt]axis cs:{ifthenelse(36.1<35.1+\distclear,36.1,35.1)},3) {35.1};
  \draw[black!85, line width=0.9pt] (axis cs:36.1,2.64) -- (axis cs:36.1,3.36);
  \addplot[draw=none, fill=negcol] coordinates {(11.5,2)};
  \node[anchor=west, font=\tiny, text=black!70, inner sep=0pt] at ([xshift=2pt]axis cs:{ifthenelse(22.8<11.5+\distclear,22.8,11.5)},2) {11.5};
  \draw[black!85, line width=0.9pt] (axis cs:22.8,1.64) -- (axis cs:22.8,2.36);
  \addplot[draw=none, fill=negcol] coordinates {(22.7,1)};
  \node[anchor=west, font=\tiny, text=black!70, inner sep=0pt] at ([xshift=2pt]axis cs:{ifthenelse(30.0<22.7+\distclear,30.0,22.7)},1) {22.7};
  \draw[black!85, line width=0.9pt] (axis cs:30.0,0.64) -- (axis cs:30.0,1.36);
\nextgroupplot[title={Claude\strut\\[-3pt]{\color{black!55}\tiny\mdseries\strut $N{=}4{,}832$}}]
  \addplot[draw=none, fill=poscol] coordinates {(92.4,5)};
  \node[anchor=east, font=\tiny, text=white, inner sep=0pt] at ([xshift=-2pt]axis cs:92.4,5) {92.4};
  \draw[black!85, line width=0.9pt] (axis cs:4.2,4.64) -- (axis cs:4.2,5.36);
  \addplot[draw=none, fill=negcol] coordinates {(5.4,4)};
  \node[anchor=west, font=\tiny, text=black!70, inner sep=0pt] at ([xshift=2pt]axis cs:{ifthenelse(6.8<5.4+\distclear,6.8,5.4)},4) {5.4};
  \draw[black!85, line width=0.9pt] (axis cs:6.8,3.64) -- (axis cs:6.8,4.36);
  \addplot[draw=none, fill=negcol] coordinates {(1.2,3)};
  \node[anchor=west, font=\tiny, text=black!70, inner sep=0pt] at ([xshift=2pt]axis cs:{ifthenelse(36.1<1.2+\distclear,36.1,1.2)},3) {1.2};
  \draw[black!85, line width=0.9pt] (axis cs:36.1,2.64) -- (axis cs:36.1,3.36);
  \addplot[draw=none, fill=negcol] coordinates {(0.2,2)};
  \node[anchor=west, font=\tiny, text=black!70, inner sep=0pt] at ([xshift=2pt]axis cs:{ifthenelse(22.8<0.2+\distclear,22.8,0.2)},2) {0.2};
  \draw[black!85, line width=0.9pt] (axis cs:22.8,1.64) -- (axis cs:22.8,2.36);
  \addplot[draw=none, fill=negcol] coordinates {(0.8,1)};
  \node[anchor=west, font=\tiny, text=black!70, inner sep=0pt] at ([xshift=2pt]axis cs:{ifthenelse(30.0<0.8+\distclear,30.0,0.8)},1) {0.8};
  \draw[black!85, line width=0.9pt] (axis cs:30.0,0.64) -- (axis cs:30.0,1.36);
\nextgroupplot[title={Gemini\strut\\[-3pt]{\color{black!55}\tiny\mdseries\strut $N{=}3{,}573$}}]
  \addplot[draw=none, fill=poscol] coordinates {(79.9,5)};
  \node[anchor=east, font=\tiny, text=white, inner sep=0pt] at ([xshift=-2pt]axis cs:79.9,5) {79.9};
  \draw[black!85, line width=0.9pt] (axis cs:4.2,4.64) -- (axis cs:4.2,5.36);
  \addplot[draw=none, fill=poscol] coordinates {(7.1,4)};
  \node[anchor=west, font=\tiny, text=black!70, inner sep=0pt] at ([xshift=2pt]axis cs:7.1,4) {7.1};
  \draw[black!85, line width=0.9pt] (axis cs:6.8,3.64) -- (axis cs:6.8,4.36);
  \addplot[draw=none, fill=negcol] coordinates {(7.2,3)};
  \node[anchor=west, font=\tiny, text=black!70, inner sep=0pt] at ([xshift=2pt]axis cs:{ifthenelse(36.1<7.2+\distclear,36.1,7.2)},3) {7.2};
  \draw[black!85, line width=0.9pt] (axis cs:36.1,2.64) -- (axis cs:36.1,3.36);
  \addplot[draw=none, fill=negcol] coordinates {(1.9,2)};
  \node[anchor=west, font=\tiny, text=black!70, inner sep=0pt] at ([xshift=2pt]axis cs:{ifthenelse(22.8<1.9+\distclear,22.8,1.9)},2) {1.9};
  \draw[black!85, line width=0.9pt] (axis cs:22.8,1.64) -- (axis cs:22.8,2.36);
  \addplot[draw=none, fill=negcol] coordinates {(3.8,1)};
  \node[anchor=west, font=\tiny, text=black!70, inner sep=0pt] at ([xshift=2pt]axis cs:{ifthenelse(30.0<3.8+\distclear,30.0,3.8)},1) {3.8};
  \draw[black!85, line width=0.9pt] (axis cs:30.0,0.64) -- (axis cs:30.0,1.36);
\nextgroupplot[title={Codex\strut\\[-3pt]{\color{black!55}\tiny\mdseries\strut $N{=}990$}}]
  \addplot[draw=none, fill=poscol] coordinates {(59.1,5)};
  \node[anchor=east, font=\tiny, text=white, inner sep=0pt] at ([xshift=-2pt]axis cs:59.1,5) {59.1};
  \draw[black!85, line width=0.9pt] (axis cs:4.2,4.64) -- (axis cs:4.2,5.36);
  \addplot[draw=none, fill=negcol] coordinates {(4.7,4)};
  \node[anchor=west, font=\tiny, text=black!70, inner sep=0pt] at ([xshift=2pt]axis cs:{ifthenelse(6.8<4.7+\distclear,6.8,4.7)},4) {4.7};
  \draw[black!85, line width=0.9pt] (axis cs:6.8,3.64) -- (axis cs:6.8,4.36);
  \addplot[draw=none, fill=negcol] coordinates {(25.7,3)};
  \node[anchor=west, font=\tiny, text=black!70, inner sep=0pt] at ([xshift=2pt]axis cs:{ifthenelse(36.1<25.7+\distclear,36.1,25.7)},3) {25.7};
  \draw[black!85, line width=0.9pt] (axis cs:36.1,2.64) -- (axis cs:36.1,3.36);
  \addplot[draw=none, fill=negcol] coordinates {(2.1,2)};
  \node[anchor=west, font=\tiny, text=black!70, inner sep=0pt] at ([xshift=2pt]axis cs:{ifthenelse(22.8<2.1+\distclear,22.8,2.1)},2) {2.1};
  \draw[black!85, line width=0.9pt] (axis cs:22.8,1.64) -- (axis cs:22.8,2.36);
  \addplot[draw=none, fill=negcol] coordinates {(8.4,1)};
  \node[anchor=west, font=\tiny, text=black!70, inner sep=0pt] at ([xshift=2pt]axis cs:{ifthenelse(30.0<8.4+\distclear,30.0,8.4)},1) {8.4};
  \draw[black!85, line width=0.9pt] (axis cs:30.0,0.64) -- (axis cs:30.0,1.36);
\end{groupplot}
\path (dist c1r1.south west) -- (dist c5r1.south east) node[midway, below=12pt, font=\small\bfseries] {Percentage of modified definitions per actor};
\begin{scope}[font=\small, text=black!70, inner sep=0pt]
\path (dist c1r1.south west) -- (dist c5r1.south east) coordinate[midway, below=35pt] (key);
\fill[poscol] ([xshift=-105pt]key) rectangle +(7pt,5pt) node[right=3pt, yshift=-2.5pt, text=black!70] {above share};
\fill[negcol] ([xshift=-40pt]key) rectangle +(7pt,5pt) node[right=3pt, yshift=-2.5pt, text=black!70] {below share};
\draw[black!85, line width=0.9pt] ([xshift=25pt, yshift=-1pt]key) -- +(0,7pt) node[right=3pt, yshift=-3.5pt, text=black!70] {all definitions};
\end{scope}
\end{tikzpicture}
  \caption{Modification share as a function of minimum distance to a \texttt{sorry} across actors. The leftmost panel shows each distance's share of all available definitions. The remaining panels show the share of each actor's modifications.}
  \label{fig:sorry-distance}
\end{figure}
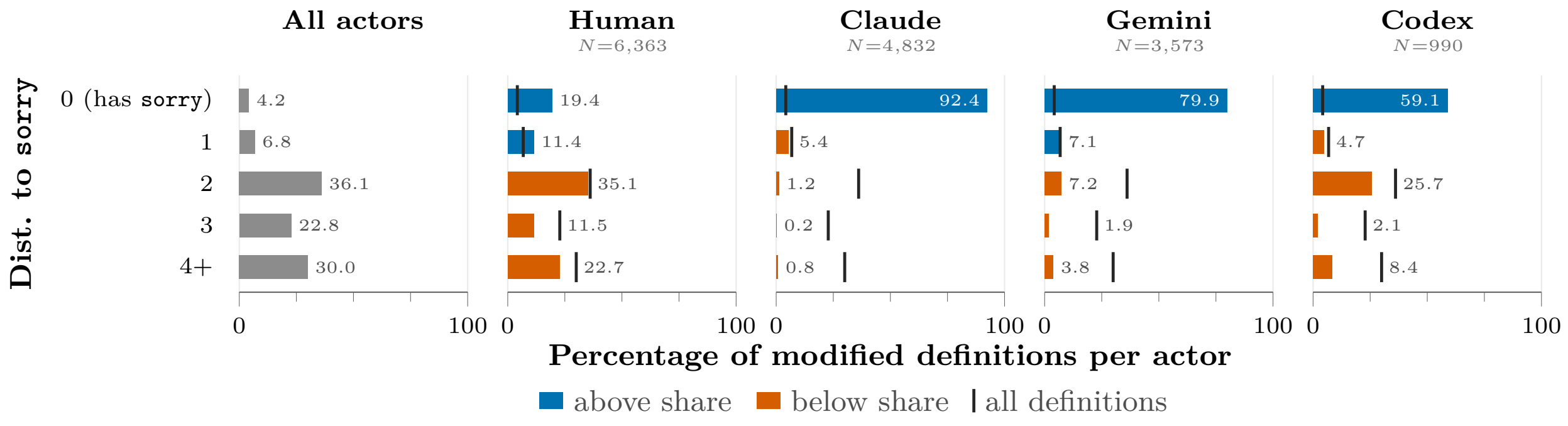

In particular, and compared to humans, we find first that AI agents {\bf focus on filling local gaps} -- AI agents are strongly attracted to editing definitions with explicit {\tt sorry} markers. This preference is clearly for the syntactic {\tt sorry} marker itself, rather than something that is associated with a problem area, as can be seen by the effect of distance from {\tt sorry}, shown in \cref{fig:sorry-distance}. All three agents show stronger-than-human preferences for editing a definition with a {\tt sorry}, and a much stronger-than-human decline in preference as one moves away from the \texttt{sorry} in the dependency graph.

In a similar vein, we find that while humans are strongly attracted to editing the longer definitions ($\beta=+2.207\pm0.055$), agent deviations from this human preference are large and negative ($-1.58$, $-1.77$ and $-2.56$ for Claude, Gemini and Codex).

As \cref{fig:coefficients} shows, there are many potential drivers of the attention of AI agents, of which focusing on \texttt{sorry} markers is only one.
Yet, regarding outcomes, working on {\tt sorry}-carrying definitions is one of the strongest signals of AI agents compared to humans. For all three agents, at least 59\% of their edits involved definitions that contained at least one {\tt sorry} (Claude, 92\%; Gemini, 80\%; Codex, 59\%), compared to just 19\% of the human edits. These preferences have knock-on effects. For example, while human edits often involve neighboring regions of the dependency graph (62\% of human edits are to definitions connected to another definition that was edited), AI agents are far more likely to make isolated edits (77\%, 69\%, 82\% for Claude, Gemini, and Codex, respectively) that do not connect to any other definitions edited in the same commit.

In short, and perhaps uncharitably, these AI agents, despite their sophistication, retain much of the flair of autocomplete systems of earlier eras. Informally, our results suggest that AI agents prefer to work on things that show basic textual features of being more obviously underdeveloped -- either relatively short, or with gaps that are explicitly marked by sorries. More often than not, they see taking the next step as a matter of attending to simple, isolated places in the code that carry clear signals that there is something to do. 

\subsubsection{Semantic Preferences}

\begin{figure}[t]
\centering
\begin{tikzpicture}
\begin{axis}[
  scale only axis, width=\linewidth-90pt, height=88pt,
  xmin=0.8, xmax=2.0, ymin=0.4, ymax=4.6,
  ytick={1,...,4}, yticklabels={{\shortstack[r]{Codex\\[-1pt]{\color{black!55}\scriptsize $N{=}990$}}},{\shortstack[r]{Gemini\\[-1pt]{\color{black!55}\scriptsize $N{=}3{,}573$}}},{\shortstack[r]{Claude\\[-1pt]{\color{black!55}\scriptsize $N{=}4{,}831$}}},{\shortstack[r]{Human\\[-1pt]{\color{black!55}\scriptsize $N{=}6{,}361$}}}},
  xtick distance=0.1, minor x tick num=1,
  xticklabel style={/pgf/number format/.cd, fixed, fixed zerofill, precision=1},
  tick label style={font=\small}, label style={font=\small\bfseries},
  xlabel={Per-token surprise (nats/token)},
  axis x line*=bottom, axis y line*=left, y axis line style={draw=none},
  tick align=outside, x axis line style={draw=black!60, -}, xtick style={black!60},
  ytick style={draw=none}, ymajorgrids, major grid style={draw=black!8},
  error bars/x dir=both, error bars/x explicit, error bars/error mark=|,
  error bars/error mark options={mark size=2pt, line width=0.6pt},
  error bars/error bar style={line width=0.6pt},
  clip=false,
]
\fill[black!8] (axis cs:0.8234,0.4) rectangle (axis cs:0.8261,4.6);
\draw[black!50, densely dashed] (axis cs:0.825,0.4) -- (axis cs:0.825,4.6)
  node[above, font=\small, text=black!70, align=center] {All definitions\\[-1pt](0.825)};
\node[anchor=south west, font=\small\bfseries, inner sep=0pt] at ([xshift=12pt]axis cs:2.0,4.6) {Change};
\addplot[negcol, mark=*, mark size=1.8pt] coordinates {(0.991,4) +- (0.0326,0)};
\addplot[negcol, mark=*, mark size=1.8pt, mark options={fill=white, line width=0.8pt}] coordinates {(0.929,4) +- (0.0285,0)};
\node[anchor=west, font=\small, text=negcol, inner sep=0pt] at ([xshift=12pt]axis cs:2.0,4) {$-0.062^{***}$};
\addplot[negcol, mark=*, mark size=1.8pt] coordinates {(1.925,3) +- (0.0654,0)};
\addplot[negcol, mark=*, mark size=1.8pt, mark options={fill=white, line width=0.8pt}] coordinates {(0.844,3) +- (0.0190,0)};
\node[anchor=west, font=\small, text=negcol, inner sep=0pt] at ([xshift=12pt]axis cs:2.0,3) {$-1.081^{***}$};
\addplot[negcol, mark=*, mark size=1.8pt] coordinates {(1.765,2) +- (0.0766,0)};
\addplot[negcol, mark=*, mark size=1.8pt, mark options={fill=white, line width=0.8pt}] coordinates {(0.952,2) +- (0.0242,0)};
\node[anchor=west, font=\small, text=negcol, inner sep=0pt] at ([xshift=12pt]axis cs:2.0,2) {$-0.813^{***}$};
\addplot[negcol, mark=*, mark size=1.8pt] coordinates {(1.479,1) +- (0.1040,0)};
\addplot[negcol, mark=*, mark size=1.8pt, mark options={fill=white, line width=0.8pt}] coordinates {(0.874,1) +- (0.0359,0)};
\node[anchor=west, font=\small, text=negcol, inner sep=0pt] at ([xshift=12pt]axis cs:2.0,1) {$-0.605^{***}$};
\begin{scope}[font=\small, text=black!70]
\draw[black!70, line width=0.8pt, fill=white] ([xshift=-62pt, yshift=8pt]axis cs:2.0,4.6) circle (1.8pt) node[right=5pt, inner sep=0pt] {after};
\fill[black!70] ([xshift=-30pt, yshift=8pt]axis cs:2.0,4.6) circle (1.8pt) node[right=5pt, inner sep=0pt] {before};
\end{scope}
\end{axis}
\end{tikzpicture}
\caption{Average per-token surprise under DeepSeek-Prover-V2-7B for definitions containing at least one \texttt{sorry} before ($\bullet$) and after ($\circ$) modification with 95\% bootstrapped intervals. The dashed line marks the mean surprise over all parent-state definitions. AI agents are systematically drawn to high-surprise bodies, and their modifications lower the surprise of those bodies. Marker: $^{***}\,p<0.001$.}
\label{fig:post-edit-surprise}
\end{figure}
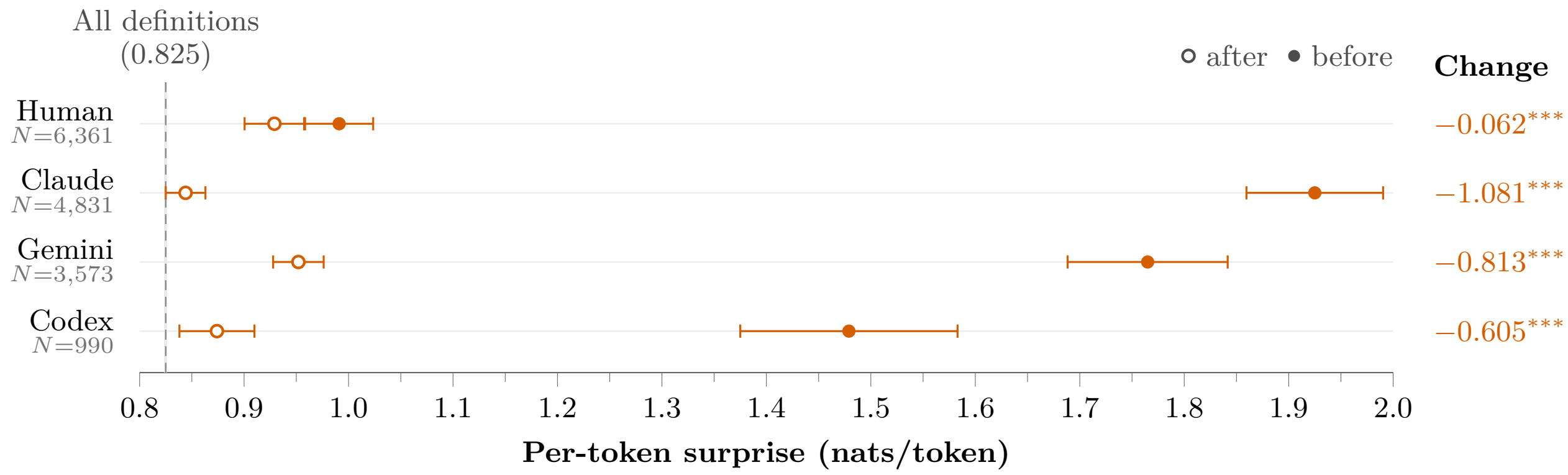

Relative to humans, AI agents may prefer shorter definitions and \texttt{sorry} markers, but as can be seen in \cref{fig:coefficients}, they also show systematic preferences for the semantic features of the repositories they are pointed at -- preferences that both mimic, and magnify, tendencies that their human parallels show. Agents do not edit \emph{all} sorries, and even the most assiduous {\tt sorry}-editor, Claude, attacks about a tenth of the sorry-bearing definitions available (10\% pooled over commits, 18\% on average per commit).

In particular, the quadratic fit on the maximum shortest distance to a leaf (Max leaf) and shortest distance to a main root (Root distance) is a proxy for how far a given definition sits from the peripheries of the dependency graph. We observe a negative coefficient on Max leaf and a positive coefficient on its square across all actors, implying a U-shaped relationship: for humans the minimum of the curve is at the lowest observed ranks (10th percentile, the leaves themselves) and the curve rises steadily with distance into the graph. In contrast, edit probability for the AI agents is at a minimum for definitions one (Codex, 38th percentile) to three (Gemini, 42nd; Claude, 52nd) dependency steps above a leaf, rising in both directions.

The second positional measure, Root distance, has the same structure. While humans show a weak preference to edit definitions close to a project's main roots, AI agents show a strong U-shaped response whose minimum sits around three steps from a main root (Gemini, 42nd percentile; Codex, 47th; Claude, 53rd), and whose curvature is significantly stronger than the human one (deviations on the squared term $+0.94$, $+0.94$ and $+4.84$ for Claude, Gemini and Codex; all $p<0.01$). These results persist in simpler regressions with either Max leaf or Root distance alone. 

Taken together, or separately, these results show that AI agents strongly prefer to {\bf modify near the logical boundaries of a proof}, either near the bottom (``axiomward''; lower percentile) or top (far from those leaves; higher percentile) of the graph. Informally, both signals amount to a preference for more peripheral engagement with the problem at hand -- AI agents prefer, far more strongly, to work backwards from the final theorem, to work forwards from the axioms, and to avoid the more densely connected parts of the argument in the middle. Humans show some of these tendencies, but at a much weaker rate; humans have weaker preferences for the extremes, and are more comfortable going into the thick of the action, working far more on the intermediate lemmas that turn more basic facts into definitions that higher-level stages of the proof rely on. 

The second main semantic difference is that all three AI agents show a strong tendency to {\bf preferentially attack high-surprise proof bodies}. The outcomes of this preference can be seen in \cref{fig:post-edit-surprise}, which reveals a critical downstream effect of this preference: AI agents tend to {\bf regularize proofs towards an LLM standard}. This effect holds when considering {\tt sorry} and non-{\tt sorry} bodies separately; see Appendix Table~\ref{tab:post-edit-by-sorry}.

In addition, all three agents also show significant relative tendencies to {\bf avoid high-surprise signatures}. Informally, the stranger the proof \emph{goal} appears, relative to an LLM standard, the less likely AI agents are to modify the body. The relative attraction to surprising bodies and aversion to surprising signatures hold regardless of which model is used to score surprise (Appendix Table~\ref{tab:surprise-sensitivity}).

\section{Discussion}

Our simulated collaboration method reveals strong and systematic deviations, with both high statistical significance and large effect sizes, from human behavior. Nearly all of these deviations are found across all three models.

These deviations come in two distinct forms, which we can broadly characterize as those that come from (1) the system's interpretation of overt {\tt sorry} signals as implicit requests for assistance and primary targets for modification, and (2) differences between the system's and the human's understanding of the overall proof structure. These edits are done in such a way that (3) agents regularize the surprise of the proof down towards the baseline.

\subsection{Agents are \texttt{sorry} autofillers}

The first class of deviations comes from how agents interpret \texttt{sorry} markers. For a human collaborator, a \texttt{sorry} is as much a placeholder as a planning device~\cite{taoPFRBlueprint2023,zhuLeanArchitect2026}, as it records a decision to \emph{defer} a proof step so that work can proceed elsewhere.
There may be several reasons to defer a definition. It can be a routine but laborious proof, it might be expected to follow from results not yet formalized~\cite{buzzardIngredientsFermat}, or it can depend on choices about the shape of the argument that are yet to be made.

Agents, by contrast, treat each \texttt{sorry} marker as an open request for assistance. We find that at least $59\%$ of every agent's modifications involve a definition that contains a \texttt{sorry}, while for humans, this fraction drops to only about a fifth of their modifications. Under a prompt as minimal as our ``take the next step'' instructions, this reading is arguably not unreasonable. As far as the Lean compiler is concerned, a complete formalization requires that no \texttt{sorry} remains, and these markers are the only explicit statement of what is left to do. Therefore, the agents adopt the most literal operationalization of progress available to them, which produces the result that agents fill the gaps that humans have chosen to defer.

Proofs are also not single-unit structures. Definitions, say, one step removed from a \texttt{sorry} are very much a part of the same unfinished argument, because they either supply the argument's inputs -- represented as outgoing edges in the dependency graph -- or use its conclusion (incoming edges). Yet, we find that agents are attracted to the \texttt{sorry} markers themselves, rather than to the regions around the proofs they mark.

In contrast, human edits are much better spread across the graph, with their largest share targeting definitions at a distance of two steps. Presumably, if collaborators understand the \texttt{sorry} as a symptom of an incomplete argument -- rather than the root cause of it -- they can reasonably be expected to work on its neighborhood, building out its argument and scaffolding, for example by restating a lemma that makes the gap easier to resolve, or by noticing that a downstream use requires a stronger statement.

We also saw that humans are more strongly drawn to editing larger definitions. In a mature formalization project, these definitions are likely to be substantial, hard-won proofs where most of the mathematical work is concentrated. By contrast, agents barely remove existing code and add things at a much lower rate, suggesting that they treat the existing codebase as fixed scaffolding to be completed rather than as a work-in-progress artifact to be revised. In addition, when agents do edit, their changes tend to be larger and more isolated from one another. We find that roughly three quarters of agent-edited definitions share no dependency with any other definition edited in the same commit, compared to about one-third for humans.

To summarize, a typical human commit tends to contain small, targeted edits to a connected region of the proof. A typical agent commit is a self-contained block of proof text dropped into a single gap -- a proof ``autofiller''. 

For collaboration, these findings can cut both ways. On the one hand, they suggest a natural division of labor that may foster human--AI complementarity~\cite{gonzalezCognitiveApproachHuman2025}: closing specific, local gaps can be tedious work that humans already defer, and agents are evidently willing to take it on. On the other hand, humans' decision to defer something is itself informative, and agents appear blind to it, which in turn can have downstream consequences.

A \texttt{sorry} may mark a definition that is expected to change, is known to be false as stated, or has an eventual proof that is meant to come from an external result not yet available, among other possibilities. Such autofilling with large, disconnected proof bodies risks producing exactly the kind of verbose and hard-to-understand artifacts that mathematicians warn against~\cite{avilaSevereMisalignmentAI2026,luLeanRefactorMultiObjective2026}. In our data, there is good evidence that much of the {\tt sorry}-filling is indeed premature; a significant fraction of the time, a declaration with a {\tt sorry} chosen for editing later evolves into a definition with a new signature -- between 25\% (Codex) and 48\% (Claude) of the time.

To promote beneficial complementarity, in other words, we ought to explore how agents can recognize whether gaps ought to be filled at all, shifting agents towards reasoning about why such gaps exist in the first place.

\subsection{Agents edit the periphery}

Broadly speaking, the second class of deviations comes from AI agents preferring to attend to the ``edges'' of a proof, near the axioms or roots, and to avoid the complex middles. There are a number of underlying drivers that might lead to these preferences.

The most parsimonious explanation is that, given our minimal prompt, such definitions constitute the low-hanging fruit of the proof process. Where \texttt{sorry}-filling provides the ready-made marker of what could ostensibly be considered a \textit{useful} next step, the dependency relationships and the type-signature complexity provide a signal of what is a \textit{feasible} next step. This interpretation, however, does not yet explain why the agents seem to consider peripheral, low-clustering nodes easy to prove.

In particular, agents may be drawn to these regions because they provide better matches to training data: shorter definitions are more likely to be closer matches to Lean code that exists elsewhere on purely combinatorial grounds. %

In addition, less-connected clusters of definitions are less likely to have complex type signatures or to call on definitions that themselves have complex signatures. This is because, compared to edges near leaf nodes, the center of the proof contains a long accumulation of decisions that combine axioms and external dependencies that ought to make it more idiosyncratic and thus more difficult for the agent to find reliable next steps to take. Near the top, meanwhile, the proof structure is expected to simplify again, as the intermediate lemmas are cashed out in favor of statements that match the final type signature.

More broadly, the asymmetry between humans and agents matters because the center of the dependency graph is where a proof's core ideas are expected to appear. It is exactly these intermediate lemmas, and the choices of how to state and prove them, that make a formal proof correspond to a \emph{particular} human proof rather than merely to \emph{some} proof~\cite{dedeoCorrespondenceProblemMathematical2026}.

Mathematicians have long held that the value of a proof lies as much in the understanding it conveys as in the certainty of its conclusion~\cite{thurstonProofProgressMathematics2006,avigadMathematiciansAgeAI2026}. An agent whose contributions cluster towards the axioms, or the top-level theorem boundaries, leaves this middle layer, and with it the work through which such understanding is built, to its human collaborators.

As in the case of {\tt sorry}-filling, such a division of labor may in fact be beneficial. A process in which agents can write foundational definitions and assemble final theorems while humans develop the intermediate arguments could be complementary and consistent with mathematicians' desire for agency and understanding~\cite{collinsHumanAgencyInitial2026}. On the other hand, such an approach would mean that the agents would not participate in the work that mathematicians find most valuable. In turn, this would mean that effective human--AI workflows would need to study and account for this asymmetry rather than assume it away.

\subsection{Agents are proof regularizers}

Our final result concerns AI agents' non-human preferences for surprise. That AI agents have a relative preference against high-surprise signatures is consistent with many of the same things that can explain their preferences for the edges and low-connectivity regions of a proof: high-surprise signatures indicate places where human decision-making has produced ideas and constructions that are likely to exceed an agent's training data, and well-designed agents ought to be expected to avoid precisely these harder problems.

As above, of course, avoiding high-surprise signatures may be a net positive for complementary collaborations; while we might hope that an agent is able to make sense of an idiosyncratic signature, it is wise for it to avoid editing a proof when it cannot. 

High-surprise bodies pose a different problem. In these cases, unusual proof styles appear to provide another signal of low-hanging fruit and agents appear to be willing to bring these unusual bodies into a more regularized form.  It is difficult, however, to see this seek-out-and-regularize instinct at the proof-body level as a net positive for human--AI workflows. The phenomenon of ``LLM homogenization'' \cite{jain2025task,burtch2026hitchhiker} has led to concerns about the emergence of a monocultural artificial hivemind \cite{jiang2026artificial}; our findings show that LLMs not only homogenize text but also appear to actively look for unusual places to smooth out. 

\subsection{Broader Lessons}

The overall picture provided by our results is one of complementarity. At both the semantic and syntactic levels, agents respond strongly to features that humans consider less important and are content, at least for a time, to ignore.

This is not necessarily a good thing for the human culture of proof-making, however. In the case of surprise regularization, for example, there are natural concerns that increasing reliance may drive the products of human--AI collaborations away from the unusual and idiosyncratic methods that may drive advances in both proof-making and Lean development.

Even in cases where complementarity suggests the potential for the peaceful coexistence of human actors and AI agents, the second-order effects are more worrying. Our results show that AI agents actively avoid assisting on definitions with particular properties -- surprising type signatures, say, or those in the center of a proof. Consider, then, ``lazy'' human mathematicians who want to shift increasing amounts of work over to an AI agent. A natural, and indeed incentivized, thing for those mathematicians to do is to learn to construct and knit their definitions together in ways that make those AI-aversive properties less common -- to make their proof style more AI-friendly -- so that the agents they recruit are more willing to contribute.

Our work suggests what such a resulting proof style, optimized for agent engagement, would look like. Definition bodies would tend to be shorter, less densely connected, and closer to leaves and roots, and both their goals and their final texts would be more predictable relative to training data baselines. While these choices might aid agents in finding ways to take the next step, they might, inadvertently, frustrate human goals of synthesis, diversity, and innovation.

\paragraph{Limitations.}

Several limitations must be considered when interpreting our results.
First, our dependency graphs come from a heuristic lexical parser rather than from the Lean compiler itself. We have validated this parser, and while it shows a high degree of reliability (Appendix~\ref{apx:parser}), the reconstruction is not exact. Even an optimal parser would, however, mix together dependencies that are immediately cognitively accessible to a human with those that come from subtle interactions at compilation time; how to track all, and only, the dependencies that are human-visible remains an open question.

Second, all nine repositories are public and predate the models we test, so their later states may appear in the agents' training data. Our work, however, still establishes a lower bound on the difference between human and AI-agent preferences because unnoticed use of one of our AI agents in the raw data should reduce the human--AI differences we observe. The differences we observe, and the resulting concerns we raise, are thus conservative. Our work also treats ``the human'' as a single actor -- but the repositories we study have between 1 and 61 contributors with likely different roles. Particular humans are expected to deviate in any number of ways, and our results here refer to the human average.

Third, our work has excluded merge commits, because a merge combines work developed separately, often by several contributors over an extended period, and so does not correspond to a single decision about the next step. The work behind a large merge commit takes place somewhat behind the scenes; we did not attempt to trace these merges back to individual forks, in part because the meaning of a commit in a private fork is far more variable, and less interpretable.

Fourth, and by necessity, our regression analyses are correlational. To establish truly causal relationships would require direct interventions on the parent state, such as inserting or removing \texttt{sorry} markers -- but such interventions would, in turn, lead to highly non-natural stimuli. Our results provide information on how human--AI collaborations might evolve, but cannot be used to simulate the evolution of a repository over long periods.

\section*{Data and Code Availability}

Our full dataset, containing the nine repositories after simulated collaborations, as well as the codebase for analyzing the resulting data is at \url{https://github.com/gyevnarb/proof-divergence}.

\section*{Generative AI Usage Statement}

The authors declare the use of generative AI in the research and writing process. According to the GAIDeT taxonomy~\cite{suchikovaGAIDeTGenerativeAI2026}, the following tasks were delegated to GAI tools under full human supervision: code generation, process automation, data collection, data cleaning, data analysis, and visualization. The GAI tools used were GPT-5.2 and GPT-6 by OpenAI, Claude Opus 4.6 by Anthropic, Gemini 3.1 Pro by Google DeepMind, Goedel-Prover-V2-32B, Kimina-Prover-72B, and DeepSeek-Prover-V2-7B. Responsibility for the final manuscript lies entirely with the authors. GAI tools are not listed as authors and do not bear responsibility for the final outcomes.

\section*{Acknowledgements}
This work was supported by Grant 63750 from the John Templeton Foundation, the Survival and Flourishing Fund, and by the Institute for Complex Social Dynamics at Carnegie Mellon University. This work used Bridges-2 at Pittsburgh Supercomputing Center through allocation MTH260032 from the Advanced Cyberinfrastructure Coordination Ecosystem: Services \& Support (ACCESS) program, which is supported by National Science Foundation grants \#2138259, \#2138286, \#2138307, \#2137603, and \#2138296.

\clearpage
\printbibliography[heading=bibintoc]

\clearpage
\appendix

\section{Methods}
\label{apx:methods}

To answer our research questions, we curate a set of intellectually significant, extended examples of human--human mathematical collaboration: nine formalization projects written in the programming language Lean~\cite{deMouraLean2021} -- listed in~\cref{tab:repos}. 
These projects are available as public, open-source repositories which give us access not just to each project's final form but to its full history through the sequence of \emph{commits} that produced it. 
The project repositories we study contain up to approximately 1,800 commits and tens to thousands of function \emph{definitions} -- such as theorems and axioms -- produced by teams of up to 61 (at the time of writing), primarily human, collaborators, many of whom are highly respected members of both the mathematics and formalization communities.

Importantly, the commit histories of most of these repositories date back several years (mean $2.22$~years). As such, they track the collaborative work of many of the field's most accomplished and dedicated practitioners, rather than autonomous AI agents which became capable of research-level mathematics in 2026. 

\begin{table}
  \centering
  \caption{Overview of the nine Lean formalization repositories we analyze.
  The column \textbf{LoC} shows the lines of code measured across all files of the final commit of the project, excluding external dependencies. The column \textbf{Editors} shows the total number of contributors to the repository. The column \textbf{Commits} counts the number of commits in the repository, and \textbf{Years} shows the age of the repository in years. All data were taken from the projects' GitHub repositories on 15 June 2026.}
  \label{tab:repos}
  \setlength{\tabcolsep}{4pt}
    \begin{tabularx}{\linewidth}{@{}l >{\raggedright\arraybackslash}X r r r r@{}}
    \toprule
    \makecell[c]{\textbf{Repo}} & \textbf{Mathematical Topic} & \makecell[r]{\textbf{LoC}} & \makecell[c]{\textbf{Editors}} & \makecell[r]{\textbf{Commits}} & \makecell[r]{\textbf{Years}} \\
    \midrule
    \textit{apap}      & Arithmetic progressions, almost periodicity~\cite{apap,apap:paper}  &  7{,}509 & 11 &  472    & 3.26 \\
    \textit{duality} & Linear-optimization duality (Farkas)~\cite{duality:paper}       &  2{,}582 &  1 &  177    & 1.91 \\
    \textit{finlin}   & Linearity of functions on finite sets~\cite{finlin:paper}       &  2{,}055 &  2 &   57    & 0.66 \\
    \textit{flt}         & Fermat's Last Theorem~\cite{flt}                                           & 28{,}031 & 61 & 1{,}614 & 2.57 \\
    \textit{ordsemi} & H\"older's theorem for ordered semigroups~\cite{ordsemi}                       &  3{,}438 &  2 &  125    & 1.88 \\
    \textit{pdl}         & Propositional Dynamic Logic~\cite{pdl,pdl:paper}                    & 23{,}873 & 13 & 1{,}041 & 2.70 \\
    \textit{pfr}         & Polynomial Freiman--Ruzsa conjecture~\cite{pfr,pfr:paper}           & 21{,}452 & 36 & 1{,}768 & 2.59 \\
    \textit{sphere}   & Sphere packing in 8 dimensions~\cite{sphere,sphere:paper}              & 22{,}303 & 25 & 1{,}074 & 2.29 \\
    \textit{zeta}       & Irrationality of $\zeta(3)$~\cite{zeta:paper}                   &  4{,}157 &  2 &  133    & 2.17 \\
    \midrule
    Mean &  --  & 12{,}822.22 & 17 & 717.89 & 2.22 \\
    \bottomrule
  \end{tabularx}
\end{table}

\subsection{Simulated Collaboration}

Our overall experimental setup is shown in~\cref{fig:experimental-setup}. 
After downloading a repository, we list -- in order from oldest to newest -- all commits on the main working branch of the repository.
From this list we filter out each commit which is either a merge -- joining different, often unrelated working branches of the repository -- or where no Lean files were modified by the human collaborators.
A commit at $t$ -- written as $H_t$, also called a child state -- in our curated dataset of repositories captures what a particular human mathematician thought and did in response to both their understanding of what ought to be done next and the parent state of the repository $H_{t-1}$.
As such, each state $H_t$ is, in practice, a commit in a Git repository, which allows an actor to inspect the full edit history of the project, while we manually remove access to all subsequent states to prevent agents from looking up the next step.
At each relevant human commit of a Lean codebase, we prompt an AI agent -- in our experiments, Claude Code with Claude Opus 4.6, Codex with GPT-5.2, and Gemini CLI with Gemini 3.1 Pro -- to produce its own discrete continuation. 
The resulting AI commit, $A_t$, is the product of whatever understanding the machine could glean of the project's needs from the combination of $H_{t-1}$ and the particular prompt we provide. 
Our goal is to analyze the differences between how a human advanced the project at time $t$ from the parent state $H_{t-1}\rightarrow H_t$, and how an AI agent did so in $H_{t-1}\rightarrow A_t$, where $A_t$ is the agent's commit.
This setup enables us to make a direct comparison between the human way of mathematical work and the machine way, in a fashion we call \emph{simulated collaboration}.

A key free variable in this method is the prompt we provide. 
Users are free to range from requesting surgically targeted interventions on a single line, to broad requests for large-scale advances. 
In this study, we use an extremely minimal prompt, \textit{``take the next step towards completing the project''}, shown in~\cref{lst:prompt}, which has the benefits of simplicity and validity regardless of project content.
We note that no stopping condition was specified: agents were allowed to run as long as they saw fit without computational budget constraints.

\begin{lstlisting}[style=promptfile,caption={General template for the ``take the next step'' prompt of our study.},label={lst:prompt}]
I am working on a Lean 4 formalization of {problem_name}. Take the next step in advancing this project by editing or adding your changes to the codebase. Summarize your changes briefly in a markdown file called CHANGES.md
\end{lstlisting}

Once the prompt is chosen, however, there are clear quantitative signals to track, which we capture with a range of features extracted from the child states. In particular, because the projects are written in Lean, the dependency relationships among definitions form a hierarchy running from axioms to final theorems. 
This structure lets us pinpoint, among other things, which stages of the arguments are more or less developed, and track mathematically meaningful differences in how humans and AI agents choose to advance the mathematics at hand. 
Over the commit histories of all nine repositories, this process generates a dataset of more than 4.7 million nodes connected by 10.8 million edges which we analyze statistically.

\subsection{The Syntax and Semantics of Lean}

\begin{figure}
    \centering

    \begin{subfigure}{0.62\textwidth}
        \centering
        \includegraphics[height=5.87cm]{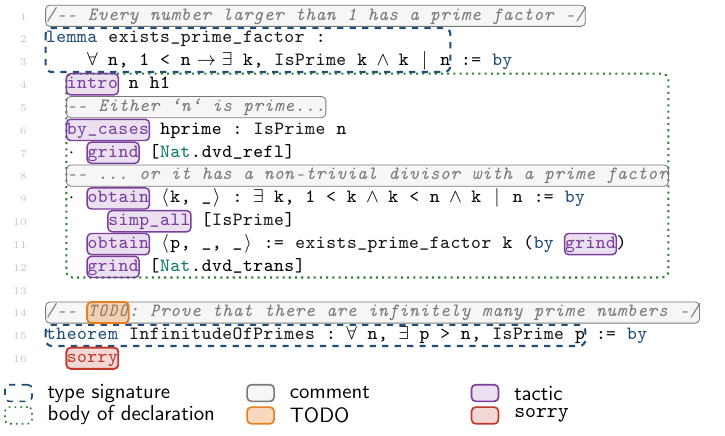}
        \caption{Example Lean~4 code highlighting the most important syntactic components of the programming language. Unfilled boxes show logical blocks, and filled boxes show keywords.}
        \label{fig:lean-code}
    \end{subfigure}
    \hfill
    \begin{subfigure}{0.37\textwidth}
        \centering
        \includegraphics[height=5.87cm]{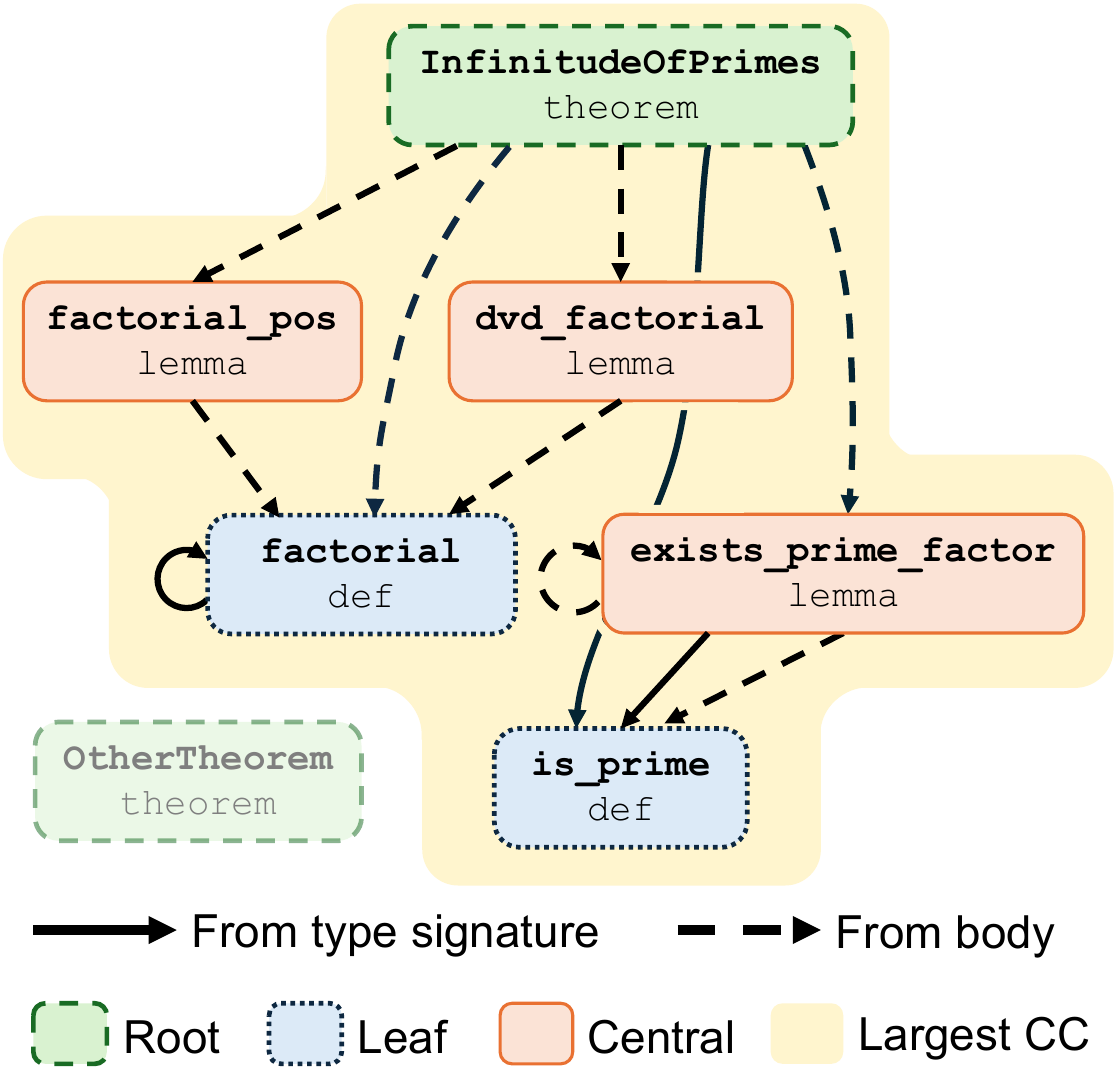}
        \caption{Nodes correspond to definitions. An arrow $A\rightarrow B$ exists if $A$ calls $B$. CC shows a (weakly) connected component.}
        \label{fig:dependency-example}
    \end{subfigure}

    \caption{Lean~4 syntax (\subref{fig:lean-code}) and dependency graph (\subref{fig:dependency-example}) based on the proof that infinitely many primes exist.}
    \label{fig:lean-4}
\end{figure}
Lean is an extensible functional programming language with a wide range of keywords and components. A proof in Lean -- a snippet of which is shown in~\cref{fig:lean-code} -- consists primarily of \emph{definitions}. These are the logical building blocks of a proof, such as basic axioms, intermediate lemmas, and high-level theorems.

For our purposes, each definition consists of two major parts: the \emph{type signature} and the \emph{body}. The type signature specifies what needs to be proven -- informally, it is the (sub)problem statement for that particular stage of the larger proof. The body provides the actual proof of the subproblem. It usually consists of (1) a sequence of \emph{tactics}, meta-programming macros that implement actions in the proof; and (2) other definitions, hopefully but not necessarily already proven, that the tactics combine in novel ways.

A crucial keyword in Lean is the \texttt{sorry} marker. Placing a \texttt{sorry} in the body of a definition amounts to an instruction to the Lean compiler to count the relevant definition (or a subpart) as already proven. This keyword then enables a human to leave one part of a proof unfinished, while making progress on later parts. Each \texttt{sorry} marker provides a clear \emph{syntactic} signal of a proof gap, and serves as our primary signal of a surface-level feature that might attract attention from either a human or an AI agent.

Part of the power of Lean lies in the ability to build on previously proven definitions and theorems, and the foundation of this power is a community-maintained library called \emph{Mathlib}~\cite{LeanMathematicalLibrary2020}. A proof in Lean is ultimately an intricately connected graph of definitions, forming a complex, directed \emph{dependency graph} between Mathlib, the proof authors' own definitions, and further potential dependencies; an example of such a dependency graph is shown in~\cref{fig:dependency-example}.

A \emph{node} in a dependency graph corresponds to any definition, while two nodes $A$, $B$ are connected by an edge $A\rightarrow B$ if definition $A$ references $B$. Definitions may reference previous ones both in their type signature and in their body. They can also call themselves recursively, meaning that dependency graphs are not necessarily directed acyclic graphs.

We call a node a \emph{leaf} if it does not reference any other definitions. Leaves are, therefore, often fundamental, low-level definitions, such as basic statements or axioms. Conversely, a \emph{root} node is not referenced by anything else, and these often correspond to high-level theorem statements and the frontier of what the project has been able to achieve. In \cref{fig:lean-code}, for example, {\tt is\_prime} is a leaf, and is referenced by {\tt dvd\_factorial} in its type signature. While definitions may sometimes be isolated from each other, a project's dependency graph is usually dominated by a large weakly \emph{connected component}. 

In contrast to the syntactic nature of the {\tt sorry} signal, the emergent network structure of a proof provides a signal of its definitions' \emph{semantics}: the network position of a definition tells us how closely it relates to other nodes -- \emph{e.g.}, is it a core part of a complex argument or a more peripheral special case -- and its position in the chain of reasoning -- \emph{e.g.}, is it towards the base of the argument, like the definition {\tt is\_prime}, or is it further along, like the use of the factorial in {\tt dvd\_factorial}.

\subsection{Operationalizing Syntactic and Semantic Development}
\label{ssec:methods:features}

We compare each human commit $H_t$ and agent commit $A_t$ with their shared human parent $H_{t-1}$. Definitions across commits are matched based on a fully qualified name that includes their declaration kind (\textit{i.e.,} theorem, lemma, def, etc.), file name, namespace, and name. We call a definition \emph{added} if its identifier is present in the child but absent from the parent, and \emph{removed} if it is present in the parent but absent from the child. We call a definition \emph{modified} if it already exists in the parent and its parsed type signature or body changes. If a definition is neither added, removed, nor modified, then it is \emph{unchanged}. We use the term \emph{edited} as a blanket term that refers to additions, modifications, and removals. We treat renaming as first a removal followed by an addition. 

We use a validated lexical parser -- see Appendix~\ref{apx:parser} for full details on parsing -- to extract the definitions and their dependency relationships from every Lean file in a commit. From this data, we extract the following features to compare human and agent proving behavior.

\paragraph{Syntactic features.}

To calculate the \emph{edit share} of an actor in a commit, we divide the number of edited definitions by the total number of definitions. We measure \emph{definition size} as the combined character count of its parsed type signature and body. To measure \emph{edit size}, we align the old and new text separately for each part using a character-level diff, then sum the characters inserted and deleted. A replacement contributes both a deletion and an insertion. A modification is \emph{body-only} when the body changes but the signature does not, and vice versa for \emph{signature-only} modifications. We count every occurrence of \texttt{sorry} markers in the parsed body. As a secondary analysis, we calculate the shortest distance from each definition to any \texttt{sorry}-carrying node, ignoring edge direction.

\paragraph{Semantic features.}

We extract features for each definition from the dependency graph of a commit as follows. First, to extract the graph itself, we create a node for each definition and an edge $A\rightarrow B$ if the parser identifies a reference to $B$ in the signature or body of $A$. Repeated references produce a single edge, and dependencies outside the repository are excluded (\emph{e.g.}, Mathlib). Such an analysis reveals a great deal about the conceptual organization of a proof, and basic graph-theoretic features of a dependency graph help us understand how, for example, modularities in graph structure enable apparently complex arguments to produce structured understanding \cite{VITERI2022105120}.

A key semantic question is \emph{where} in the proof the actor (human, or AI) chooses to intervene. We measure this in two ways. First, the \emph{maximum shortest distance to a leaf} (``max leaf'') is the longest path among all shortest paths to reachable nodes with no outgoing edges. In rare cases, there is a dependency cycle in a leaf position (\emph{i.e.}, a strongly connected set of nodes with no outgoing edges); we compute the shortest distance to any node in such terminal cycles. This measure is zero for a leaf node or terminal cycle. 

Max leaf measures how far into a proof, from the axiom level, the actor chooses to intervene. We can also measure the converse---how far from the ``theorem'' point an actor chooses to intervene. ``Theorem'' here is in quotes because there may be many top-level definitions in a proof in development; points which currently are not used as dependencies by anything else, but represent the furthest along the proof has developed. There are also (we find) a large number of nodes that nothing else depends on but that correspond to abandoned or otherwise uninteresting lines of development; these nodes are roots (\emph{i.e.}, they are not called by any other definitions in the repository) but do not, on inspection, correspond to the frontier of the proof's development.

In order to identify the theorem-like definitions among the roots, we first measure the transitive closure of each root node in the largest connected component -- \emph{i.e.}, all the nodes that that root node calls upon, or that are called upon by those nodes, and so on. We begin with the root node with the largest transitive closure, and add, in a greedy fashion, additional root nodes until the total size of the transitive closure of all the root nodes so far increases by less than 5\%. This heuristic allows us to capture the set of ``main'' root nodes, while dropping the roots that are simply by-products of testing and experimentation.

Having identified the main root nodes, we can then compute the \emph{undirected shortest distance to a main root} (``Root distance''). This distance is undirected because in some cases (\emph{e.g.}, a root that was not selected as a main root on the basis of closure) reaching a main root requires going, for a small number of steps, in the direction of dependencies. Root distance and max leaf are, naturally, anticorrelated ($-0.23$), but not perfectly so, because of how the graph can widen as one moves from axiom to theorem. We include both in our analysis.

Finally, we measure the per-token \emph{surprise} of both a definition's body and its type signature. This is computed using a language model fine-tuned for proof-completion tasks with Lean 4. Per-token surprise is calculated for non-comment tokens only and using, for context, the full contents of the file so far; in rare cases, a file is longer than the model's context window. These longer files are scored in chunks of whole declarations of at most 24,000 tokens, each preceded by up to 12,000 tokens of the file as context.

If, for example, the type signature in question is located between tokens $k_\mathrm{start}$ and $k_\mathrm{end}$ (inclusive), then the surprise is
\begin{equation}
    S = -\frac{1}{k_\mathrm{end}-k_\mathrm{start}+1} \sum_{i=k_\mathrm{start}}^{k_\mathrm{end}} \log{P(t_i \mid t_{0:(i-1)})},
\end{equation}
where $t_i$ is the $i$th token and $t_{0:(i-1)}$ is the list of the preceding context tokens. High surprise means that the tokens are more improbable under the model; the core probability, $P(t_i \mid t_{0:(i-1)})$, is computed using three models, DeepSeek-Prover-V2-7B, Kimina-Prover-72B, and Goedel-Prover-V2-32B. Goedel and Kimina are based on the Qwen architecture, while DeepSeek-Prover is based on the LLaMA architecture. All three models lead to essentially the same conclusions; DeepSeek-Prover-V2-7B provides the best fit under AIC for our main regression, and so we present its results in the main text.

When used in a regression model, all of these continuous predictors are converted to ranks computed once per parent commit over the definitions in the largest weakly connected component. Regression models use predictors based on features of the parent graph restricted to definitions in its largest weakly connected component.

\subsection{Statistical Methods and Regression Analysis}
\label{ssec:methods:stats}

Our analysis focuses on identifying which definitions are chosen for modification by the different actors. We first characterize the related behavior of actors using descriptive statistics and comparisons. Unless otherwise noted, all $p$-values reported are based on a beta-binomial model which controls for differences between commits and between repositories, and allows the difference in edit shares between actors to vary by repository. Pairwise differences are tested with Wald tests and we use Holm correction to account for multiple testing.

We also seek to understand the main driving factors behind the observed modification behaviors, which we model using logistic regression. At a high level, we fit a binomial logistic mixed-effects model using the R programming language, specifically the \texttt{glmmTMB} package. Our model predicts whether a parent-state definition is modified in the child state.  We restrict the input data to definitions in the largest connected component of the shared parent graph, as otherwise structural measures may be biased unnecessarily. Overall, this restriction removes approximately $12.2\%$ of definitions across all repositories and commits.

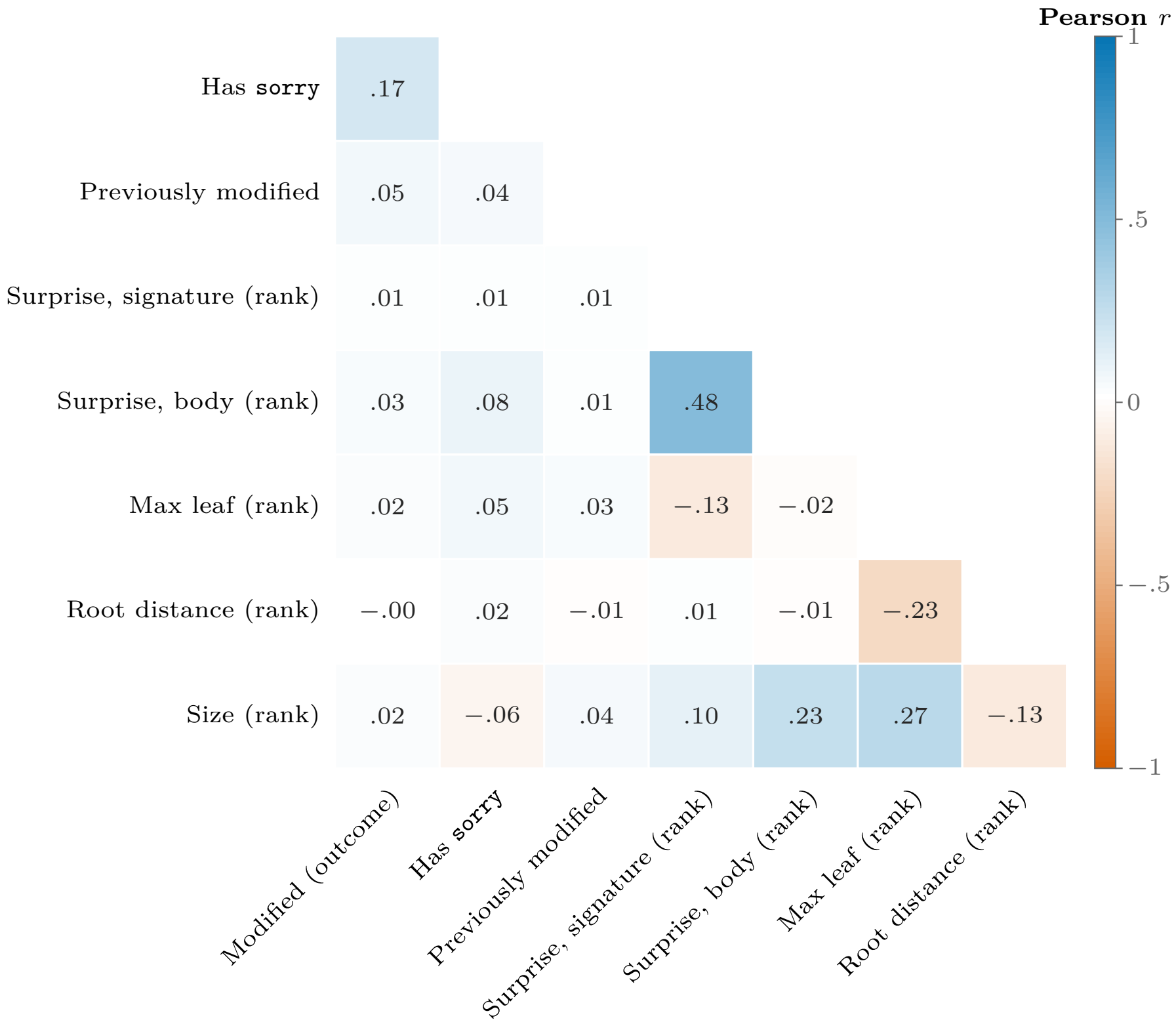
\begin{figure}
  \centering
  \setbox0\hbox{\scriptsize\begin{tabular}{@{}l@{}}Modified (outcome)\\Has \texttt{sorry}\\Previously modified\\Surprise, signature (rank)\\Surprise, body (rank)\\Max leaf (rank)\\Root distance (rank)\\Size (rank)\end{tabular}}\edef\corlabelw{\the\wd0}%
\pgfmathsetlengthmacro\corcell{min(30pt, (\linewidth - \corlabelw - 39pt) / 7)}
\ifdim\corcell<20pt\def\corvalfont{\tiny}\else\def\corvalfont{\scriptsize}\fi
\begin{tikzpicture}[x=\corcell, y=\corcell, font=\scriptsize, inner sep=1pt]
\fill[poscol!17!white, draw=white, line width=0.6pt] (0,-1) rectangle (1,0) node[midway, font=\corvalfont, text=black!85] {.17};
\fill[poscol!5!white, draw=white, line width=0.6pt] (0,-2) rectangle (1,-1) node[midway, font=\corvalfont, text=black!85] {.05};
\fill[poscol!4!white, draw=white, line width=0.6pt] (1,-2) rectangle (2,-1) node[midway, font=\corvalfont, text=black!85] {.04};
\fill[poscol!1!white, draw=white, line width=0.6pt] (0,-3) rectangle (1,-2) node[midway, font=\corvalfont, text=black!85] {.01};
\fill[poscol!1!white, draw=white, line width=0.6pt] (1,-3) rectangle (2,-2) node[midway, font=\corvalfont, text=black!85] {.01};
\fill[poscol!1!white, draw=white, line width=0.6pt] (2,-3) rectangle (3,-2) node[midway, font=\corvalfont, text=black!85] {.01};
\fill[poscol!3!white, draw=white, line width=0.6pt] (0,-4) rectangle (1,-3) node[midway, font=\corvalfont, text=black!85] {.03};
\fill[poscol!8!white, draw=white, line width=0.6pt] (1,-4) rectangle (2,-3) node[midway, font=\corvalfont, text=black!85] {.08};
\fill[poscol!1!white, draw=white, line width=0.6pt] (2,-4) rectangle (3,-3) node[midway, font=\corvalfont, text=black!85] {.01};
\fill[poscol!48!white, draw=white, line width=0.6pt] (3,-4) rectangle (4,-3) node[midway, font=\corvalfont, text=black!85] {.48};
\fill[poscol!2!white, draw=white, line width=0.6pt] (0,-5) rectangle (1,-4) node[midway, font=\corvalfont, text=black!85] {.02};
\fill[poscol!5!white, draw=white, line width=0.6pt] (1,-5) rectangle (2,-4) node[midway, font=\corvalfont, text=black!85] {.05};
\fill[poscol!3!white, draw=white, line width=0.6pt] (2,-5) rectangle (3,-4) node[midway, font=\corvalfont, text=black!85] {.03};
\fill[negcol!13!white, draw=white, line width=0.6pt] (3,-5) rectangle (4,-4) node[midway, font=\corvalfont, text=black!85] {$-$.13};
\fill[negcol!2!white, draw=white, line width=0.6pt] (4,-5) rectangle (5,-4) node[midway, font=\corvalfont, text=black!85] {$-$.02};
\fill[negcol!0!white, draw=white, line width=0.6pt] (0,-6) rectangle (1,-5) node[midway, font=\corvalfont, text=black!85] {$-$.00};
\fill[poscol!2!white, draw=white, line width=0.6pt] (1,-6) rectangle (2,-5) node[midway, font=\corvalfont, text=black!85] {.02};
\fill[negcol!1!white, draw=white, line width=0.6pt] (2,-6) rectangle (3,-5) node[midway, font=\corvalfont, text=black!85] {$-$.01};
\fill[poscol!1!white, draw=white, line width=0.6pt] (3,-6) rectangle (4,-5) node[midway, font=\corvalfont, text=black!85] {.01};
\fill[negcol!1!white, draw=white, line width=0.6pt] (4,-6) rectangle (5,-5) node[midway, font=\corvalfont, text=black!85] {$-$.01};
\fill[negcol!23!white, draw=white, line width=0.6pt] (5,-6) rectangle (6,-5) node[midway, font=\corvalfont, text=black!85] {$-$.23};
\fill[poscol!2!white, draw=white, line width=0.6pt] (0,-7) rectangle (1,-6) node[midway, font=\corvalfont, text=black!85] {.02};
\fill[negcol!6!white, draw=white, line width=0.6pt] (1,-7) rectangle (2,-6) node[midway, font=\corvalfont, text=black!85] {$-$.06};
\fill[poscol!4!white, draw=white, line width=0.6pt] (2,-7) rectangle (3,-6) node[midway, font=\corvalfont, text=black!85] {.04};
\fill[poscol!10!white, draw=white, line width=0.6pt] (3,-7) rectangle (4,-6) node[midway, font=\corvalfont, text=black!85] {.10};
\fill[poscol!23!white, draw=white, line width=0.6pt] (4,-7) rectangle (5,-6) node[midway, font=\corvalfont, text=black!85] {.23};
\fill[poscol!27!white, draw=white, line width=0.6pt] (5,-7) rectangle (6,-6) node[midway, font=\corvalfont, text=black!85] {.27};
\fill[negcol!13!white, draw=white, line width=0.6pt] (6,-7) rectangle (7,-6) node[midway, font=\corvalfont, text=black!85] {$-$.13};
\node[anchor=east] at ([xshift=-3pt]0,-0.5) {Has \texttt{sorry}};
\node[anchor=east] at ([xshift=-3pt]0,-1.5) {Previously modified};
\node[anchor=east] at ([xshift=-3pt]0,-2.5) {Surprise, signature (rank)};
\node[anchor=east] at ([xshift=-3pt]0,-3.5) {Surprise, body (rank)};
\node[anchor=east] at ([xshift=-3pt]0,-4.5) {Max leaf (rank)};
\node[anchor=east] at ([xshift=-3pt]0,-5.5) {Root distance (rank)};
\node[anchor=east] at ([xshift=-3pt]0,-6.5) {Size (rank)};
\node[anchor=north east, rotate=45] at ([yshift=-3pt]0.5,-7) {Modified (outcome)};
\node[anchor=north east, rotate=45] at ([yshift=-3pt]1.5,-7) {Has \texttt{sorry}};
\node[anchor=north east, rotate=45] at ([yshift=-3pt]2.5,-7) {Previously modified};
\node[anchor=north east, rotate=45] at ([yshift=-3pt]3.5,-7) {Surprise, signature (rank)};
\node[anchor=north east, rotate=45] at ([yshift=-3pt]4.5,-7) {Surprise, body (rank)};
\node[anchor=north east, rotate=45] at ([yshift=-3pt]5.5,-7) {Max leaf (rank)};
\node[anchor=north east, rotate=45] at ([yshift=-3pt]6.5,-7) {Root distance (rank)};
\shade[bottom color=negcol, top color=poscol, middle color=white] ([xshift=8pt]7,0) rectangle ([xshift=14pt]7,-7);
\draw[black!60] ([xshift=8pt]7,0) rectangle ([xshift=14pt]7,-7);
\draw[black!60] ([xshift=14pt]7,-7.000) -- ++(2pt,0) node[right=1pt, inner sep=0pt] {$-1$};
\draw[black!60] ([xshift=14pt]7,-5.250) -- ++(2pt,0) node[right=1pt, inner sep=0pt] {$-.5$};
\draw[black!60] ([xshift=14pt]7,-3.500) -- ++(2pt,0) node[right=1pt, inner sep=0pt] {0};
\draw[black!60] ([xshift=14pt]7,-1.750) -- ++(2pt,0) node[right=1pt, inner sep=0pt] {.5};
\draw[black!60] ([xshift=14pt]7,-0.000) -- ++(2pt,0) node[right=1pt, inner sep=0pt] {1};
\node[anchor=south, font=\scriptsize\bfseries] at ([xshift=11pt, yshift=2pt]7,0) {Pearson $r$};
\end{tikzpicture}
  \caption{Pairwise Pearson correlations between our (rank-normalized) syntactic and semantic features, and the modification outcome, pooled across actors and repositories.}
  \label{fig:correlations}
\end{figure}

Our fixed-effects predictors are features calculated from the parent state: the presence of \texttt{sorry} markers, Max leaf and Root distance and their squares, surprise over the type signature and the body, definition size, and an indicator of previous modification. We choose these features for their interpretability; they are largely decorrelated with each other (see~\cref{fig:correlations}) and the VIFs range from $1.00$ (previously modified) to $1.37$ (body surprise).

Every predictor interacts with each actor, allowing both the intercept and the coefficients to vary by actor. We take humans as the reference category, so an agent's coefficients are represented as deviations from the corresponding human coefficients. Finally, our random effects comprise actor-specific intercepts within repositories and a shared intercept for each repository--commit block.
Using R notation, our model is succinctly expressed as:
\begin{verbatim}
    is_modified ~
      (has_sorry + p_is_modified + 
       max_leaf_r + I(max_leaf_r^2) +
       max_root_r + I(max_root_r^2) +
       surprise_type_r + surprise_body_r + 
       length_r) * Actor +
      diag(0 + Actor | repo) + (1 | repo:commit)
\end{verbatim}

\section{Lexical parsing of Lean files}
\label{apx:parser}

\begin{table}[!ht]
\centering
\caption{Our parser's accuracy compared against the Lean~4 compiler ground truth. $N$ shows the number of definitions or dependency edges. $F_1$ measures the recovery of named definitions or the edges between them. $P$ and $R$ measure the precision and recall of the recovery of edges between definitions. $\rho_{\mathrm{out}}$/$\rho_{\mathrm{in}}$ are the Spearman rank-correlations between our parser's and the compiler's dependency graph out/in-degrees, $\rho_{\mathrm{leaf}}$/$\rho_{\mathrm{root}}$ the correlations with Max leaf and Root distance, and LCC the fraction of the largest weakly connected component recovered.}
\label{tab:validation}
\resizebox{\textwidth}{!}{%
\begin{tabular}{lrrrrrrrrrrrr}
\toprule
& & \multicolumn{2}{c}{Definitions} & \multicolumn{4}{c}{Dependency edges} & \multicolumn{2}{c}{Degree rank corr.} & \multicolumn{3}{c}{Graph features} \\
\cmidrule(lr){3-4}\cmidrule(lr){5-8}\cmidrule(lr){9-10}\cmidrule(lr){11-13}
Repository & Modules & $N$ & $F_1$ & $N$ & $P$ & $R$ & $F_1$ & $\rho_{\mathrm{out}}$ & $\rho_{\mathrm{in}}$ & $\rho_{\mathrm{leaf}}$ & $\rho_{\mathrm{root}}$ & LCC \% \\
\midrule
\textit{apap} & 40/40 & 835 & 0.998 & 2{,}173 & 0.812 & 0.830 & 0.821 & 0.79 & 0.84 & 0.70 & 0.20 & 91 \\
\textit{duality} & 7/7 & 184 & 1.000 & 951 & 0.979 & 0.432 & 0.600 & 0.80 & 0.57 & 0.62 & 0.30 & 99 \\
\textit{finlin} & 2/2 & 15 & 1.000 & 25 & 1.000 & 1.000 & 1.000 & 1.00 & 1.00 & 1.00 & 1.00 & 100 \\
\textit{flt} & 171/174 & 1{,}845 & 1.000 & 5{,}117 & 0.715 & 0.532 & 0.610 & 0.76 & 0.64 & 0.50 & 0.32 & 80 \\
\textit{ordsemi} & 16/16 & 228 & 1.000 & 1{,}122 & 0.986 & 0.608 & 0.752 & 0.84 & 0.88 & 0.84 & 0.67 & 98 \\
\textit{pdl} & 29/34 & 967 & 1.000 & 5{,}890 & 0.916 & 0.429 & 0.585 & 0.75 & 0.77 & 0.69 & 0.65 & 96 \\
\textit{pfr} & 71/71 & 1{,}010 & 0.997 & 4{,}383 & 0.884 & 0.710 & 0.787 & 0.93 & 0.77 & 0.92 & 0.43 & 94 \\
\textit{sphere} & 88/88 & 1{,}614 & 0.999 & 4{,}700 & 0.936 & 0.761 & 0.839 & 0.85 & 0.85 & 0.86 & 0.60 & 84 \\
\textit{zeta} & 8/8 & 131 & 1.000 & 287 & 0.948 & 0.951 & 0.950 & 0.97 & 0.96 & 0.76 & 0.92 & 98 \\
\midrule
Pooled & 432/440 & 6{,}829 & 0.999 & 24{,}648 & 0.862 & 0.614 & 0.717 & --- & --- & --- & --- & --- \\
\bottomrule
\end{tabular}%
}
\end{table}

While a simple dependency graph as in \cref{fig:dependency-example} can be easily read off by human eyes, the projects we consider grow to enormous complexity and Lean's metaprogramming abilities mean that the only way to establish this graph with certainty is to compile the code in Lean's own parsed environments. This is, unfortunately, not possible at the scale required, for two reasons.

First, obtaining a built Lean environment requires that the project compile and be error-free, but this is rarely the case for either the work-in-progress repositories or the changes made by AI agents. Second, even if the project is error-free, doing so for many tens of thousands of commit stages is prohibitive in both time and compute; builds must be pegged to a particular version of both Lean and Mathlib, and can require minutes to hours of compilation time and many gigabytes of storage.

Therefore, we use our own lexical parser, based on a combination of regular expressions and heuristics, which is scalable and applies uniformly to every commit whether or not it compiles. This parser is approximate and its use may introduce measurement errors. To quantify this error, we validate its outputs against ground truth taken from the Lean~4 compiler itself as follows. 

For each of the nine repositories, we build the full project at its final commit and load the resulting environment.  This environment provides the ground truth record of what each definition depends on. We restrict attention to repository-internal targets, since references to \textit{Mathlib} or the core library -- between $82\%$ and $99\%$ of all references -- lie outside the graph we study in our experiments. We report precision, recall, and $F_1$ at two levels: the \emph{definitions} and \emph{dependency edges}. The first level asks whether the lexical parser recovers the same set of named objects as the Lean environment. The second level asks whether the parser recovers the same dependencies between the identified definitions. Because our analyses rest on graph-structural features rather than on individual edges, we also report Spearman rank correlations between the parser and the compiler out- and in-degrees, written as $\rho_{\mathrm{out}}$ and $\rho_{\mathrm{in}}$.

Our parser is essentially exact at identifying definitions ($F_1 = 0.999$ over $6{,}829$ declarations), so the nodes of our graphs are reliable. Edge recovery has high precision and moderate recall ($P = 0.862$, $R = 0.614$): the dependencies we report are almost always real, but the parser systematically under-counts them. 

The two position measures the regression uses are (mostly) reproduced well. The max leaf distance is recovered well ($\rho_{\mathrm{leaf}} = 0.50$ to $1.00$ across projects; $0.50$ to $0.92$ in the four largest), as is membership of the giant component ($80$ to $100\%$ of declarations). The root distance is noisier ($\rho_{\mathrm{root}} = 0.20$ to $1.00$; $0.32$ to $0.65$ in the four largest). Root distance can vary from ground truth for two reasons: in the selection of the ``main'' root set, and in the computation of the minimum paths.

Much of the shortfall in recall is intrinsic to the way Lean~4 works. By a token-matching heuristic, roughly a quarter ($24\%$) of the dependencies identified by the kernel are never written by name in the declaration's text but introduced during elaboration (instance synthesis, variable binders, coercions), and a further $2.6\%$ are expressed only through notation declared in the repository; among the dependencies that are written by name, recall is $0.84$. Given the goals of this work -- an analysis of how humans and AI agents work with the source text of a repository -- the written dependency structure is the relevant target, but we do not treat this heuristic as a bound on what a lexical parser could recover.

\section{Additional Tables}
\label{apx:tables}

\Cref{tab:surprise-sensitivity,tab:post-edit-by-sorry,tab:coefficients_full} report further results and exact numbers underlying our figures in~\cref{sec:results}.

\begin{table}[ht]
\centering
\caption{Sensitivity of the coefficients for surprise in Figure~\ref{fig:coefficients} to the language model used to score surprise. Each column refits the same model, but with surprise ranks computed under different LLMs. Log-likelihoods are comparable across columns.}
\label{tab:surprise-sensitivity}
\resizebox{\columnwidth}{!}{%
\begin{tabular}{@{}llccc@{}}
\toprule
\textbf{Predictor} & & \textbf{DeepSeek-Prover-V2-7B} & \textbf{Goedel-Prover-V2-32B} & \textbf{Kimina-Prover-72B} \\
\midrule
Surprise, signature & Human & $\phantom{-}0.024\,(0.056)$ & $\phantom{-}0.087\,(0.055)$ & $\phantom{-}0.107^{*}\,(0.054)$ \\
 & $\Delta$ Claude & ${\color{negcol}-0.877^{***}\,(0.085)}$ & ${\color{negcol}-0.528^{***}\,(0.083)}$ & ${\color{negcol}-0.531^{***}\,(0.082)}$ \\
 & $\Delta$ Gemini & ${\color{negcol}-0.338^{***}\,(0.089)}$ & $-0.156\,(0.089)$ & ${\color{negcol}-0.191^{*}\,(0.087)}$ \\
 & $\Delta$ Codex & ${\color{negcol}-0.390^{**}\,(0.138)}$ & ${\color{negcol}-0.528^{***}\,(0.139)}$ & ${\color{negcol}-0.498^{***}\,(0.137)}$ \\
Surprise, body & Human & $\phantom{-}0.532^{***}\,(0.058)$ & $\phantom{-}0.293^{***}\,(0.058)$ & $\phantom{-}0.552^{***}\,(0.056)$ \\
 & $\Delta$ Claude & ${\color{poscol}\phantom{-}1.029^{***}\,(0.084)}$ & ${\color{poscol}\phantom{-}0.995^{***}\,(0.080)}$ & ${\color{poscol}\phantom{-}0.672^{***}\,(0.079)}$ \\
 & $\Delta$ Gemini & ${\color{poscol}\phantom{-}0.854^{***}\,(0.091)}$ & ${\color{poscol}\phantom{-}1.064^{***}\,(0.086)}$ & ${\color{poscol}\phantom{-}0.831^{***}\,(0.086)}$ \\
 & $\Delta$ Codex & ${\color{poscol}\phantom{-}0.522^{***}\,(0.135)}$ & ${\color{poscol}\phantom{-}0.955^{***}\,(0.133)}$ & ${\color{poscol}\phantom{-}0.349^{**}\,(0.133)}$ \\
\midrule
Log-likelihood & & $-64{,}048$ & $-64{,}116$ & $-64{,}124$ \\
AIC & & $128{,}185$ & $128{,}323$ & $128{,}339$ \\
\bottomrule
\end{tabular}
}%
\end{table}

\begin{table}[ht]
\centering
\caption{Change in surprise over definition body after a modification (nats per token), split by whether the definition contained a \texttt{sorry} in the parent commit, and by each language model used to score surprise. Errors are cluster-bootstrapped standard errors over commits, with the number of definitions shown in parentheses.}
\label{tab:post-edit-by-sorry}
\resizebox{\textwidth}{!}{%
\begin{tabular}{@{}llccc@{}}
\toprule
\textbf{Actor} & \textbf{Parent body} & \textbf{Goedel-Prover-V2-32B} & \textbf{Kimina-Prover-72B} & \textbf{DeepSeek-Prover-V2-7B} \\
\midrule
Human & with \texttt{sorry} & $-0.063 \pm 0.023^{**}$ (1,232) & $-0.089 \pm 0.023^{***}$ (1,232) & $-0.345 \pm 0.034^{***}$ (1,232) \\
 & without \texttt{sorry} & $+0.008 \pm 0.007$ (5,129) & $-0.007 \pm 0.009$ (5,129) & $+0.006 \pm 0.007$ (5,129) \\
Claude & with \texttt{sorry} & $-0.401 \pm 0.022^{***}$ (4,464) & $-0.468 \pm 0.023^{***}$ (4,464) & $-1.157 \pm 0.039^{***}$ (4,464) \\
 & without \texttt{sorry} & $-0.187 \pm 0.024^{***}$ (367) & $-0.188 \pm 0.028^{***}$ (367) & $-0.157 \pm 0.022^{***}$ (367) \\
Gemini & with \texttt{sorry} & $-0.372 \pm 0.026^{***}$ (2,855) & $-0.445 \pm 0.030^{***}$ (2,855) & $-0.994 \pm 0.046^{***}$ (2,855) \\
 & without \texttt{sorry} & $-0.109 \pm 0.020^{***}$ (718) & $-0.120 \pm 0.021^{***}$ (718) & $-0.092 \pm 0.019^{***}$ (718) \\
Codex & with \texttt{sorry} & $-0.442 \pm 0.063^{***}$ (585) & $-0.441 \pm 0.070^{***}$ (585) & $-0.984 \pm 0.089^{***}$ (585) \\
 & without \texttt{sorry} & $-0.113 \pm 0.015^{***}$ (405) & $-0.107 \pm 0.018^{***}$ (405) & $-0.058 \pm 0.015^{***}$ (405) \\
\bottomrule
\end{tabular}%
}
\end{table}

\begin{table}[!ht]
\centering
\caption{Fixed-effect coefficients corresponding to \cref{fig:coefficients}, main text.}
\label{tab:coefficients_full}
\setlength{\tabcolsep}{6pt}
\resizebox{\columnwidth}{!}{%
\begin{tabular}{@{}lcccc@{}}
\toprule
\textbf{Predictor} & \textbf{Human} & $\Delta$\,\textbf{Claude} & $\Delta$\,\textbf{Gemini} & $\Delta$\,\textbf{Codex} \\
\midrule
Intercept & $-6.612^{***}\,(0.354)$ & ${\color{negcol}-1.119^{*\phantom{**}}\,(0.546)}$ & ${\color{negcol}-1.185^{*\phantom{**}}\,(0.546)}$ & $-0.364^{\phantom{***}}\,(0.644)$ \\
Has \texttt{sorry} & $\phantom{-}1.969^{***}\,(0.037)$ & ${\color{poscol}\phantom{-}4.556^{***}\,(0.071)}$ & ${\color{poscol}\phantom{-}3.074^{***}\,(0.060)}$ & ${\color{poscol}\phantom{-}1.714^{***}\,(0.084)}$ \\
Surprise, signature (rank) & $\phantom{-}0.024^{\phantom{***}}\,(0.056)$ & ${\color{negcol}-0.877^{***}\,(0.085)}$ & ${\color{negcol}-0.338^{***}\,(0.089)}$ & ${\color{negcol}-0.390^{**\phantom{*}}\,(0.138)}$ \\
Surprise, body (rank) & $\phantom{-}0.532^{***}\,(0.058)$ & ${\color{poscol}\phantom{-}1.029^{***}\,(0.084)}$ & ${\color{poscol}\phantom{-}0.854^{***}\,(0.091)}$ & ${\color{poscol}\phantom{-}0.522^{***}\,(0.135)}$ \\
Max leaf (rank) & $-0.125^{\phantom{***}}\,(0.215)$ & ${\color{negcol}-4.745^{***}\,(0.329)}$ & ${\color{negcol}-1.584^{***}\,(0.356)}$ & ${\color{negcol}-3.533^{***}\,(0.549)}$ \\
Max leaf (rank)$^{2}$ & $\phantom{-}0.611^{**\phantom{*}}\,(0.189)$ & ${\color{poscol}\phantom{-}4.054^{***}\,(0.296)}$ & ${\color{poscol}\phantom{-}1.424^{***}\,(0.315)}$ & ${\color{poscol}\phantom{-}4.223^{***}\,(0.481)}$ \\
Root distance (rank) & $-0.939^{***}\,(0.193)$ & ${\color{negcol}-0.750^{*\phantom{**}}\,(0.312)}$ & $-0.405^{\phantom{***}}\,(0.335)$ & ${\color{negcol}-4.204^{***}\,(0.496)}$ \\
Root distance (rank)$^{2}$ & $\phantom{-}0.648^{***}\,(0.189)$ & ${\color{poscol}\phantom{-}0.938^{**\phantom{*}}\,(0.297)}$ & ${\color{poscol}\phantom{-}0.945^{**\phantom{*}}\,(0.315)}$ & ${\color{poscol}\phantom{-}4.841^{***}\,(0.463)}$ \\
Size (rank) & $\phantom{-}2.207^{***}\,(0.055)$ & ${\color{negcol}-1.579^{***}\,(0.082)}$ & ${\color{negcol}-1.766^{***}\,(0.086)}$ & ${\color{negcol}-2.562^{***}\,(0.130)}$ \\
Previously modified & $\phantom{-}1.947^{***}\,(0.048)$ & ${\color{negcol}-0.745^{***}\,(0.080)}$ & ${\color{negcol}-1.254^{***}\,(0.090)}$ & ${\color{negcol}-1.255^{***}\,(0.143)}$ \\
\midrule
$N$ & $1{,}025{,}878$ & $1{,}027{,}626$ & $1{,}028{,}055$ & $1{,}028{,}671$ \\
$N$ (edited) & $6{,}356$ ($0.62\%$) & $4{,}830$ ($0.47\%$) & $3{,}572$ ($0.35\%$) & $989$ ($0.10\%$) \\
AUC (fixed effects) & $0.742$ & $0.968$ & $0.905$ & $0.802$ \\
AUC (with random effects) & $0.924$ & $0.974$ & $0.950$ & $0.876$ \\
\bottomrule
\end{tabular}%
}
\end{table}

\end{document}